%% file: ms.tex
\documentclass[lettersize,journal]{IEEEtran}

\IEEEoverridecommandlockouts                              

\input{preamble.tex}

\title{\LARGE \bf
{\sc \textbf{Cafe-Mpc}}: A Cascaded-Fidelity Model Predictive Control Framework with Tuning-Free Whole-Body Control
}

\author{He Li and Patrick M. Wensing 
\thanks{*This work was supported in part by a research gift from Google and through NSF Grant CMMI-2220924 and ONR Award N0001420WX01278 via subawards to the University of Notre Dame.}
\thanks{He Li and Patrick M. Wensing are with Department of Aerospace and Mechanical Engineering,
        University of Notre Dame, Notre Dame, IN 46556 USA
        ({\tt\small hli25@nd.edu, pwensing@nd.edu})}%
}

\begin{document}

\maketitle
\thispagestyle{plain}
\pagestyle{plain}

\input{MS_01/00_Abstract}
\input{MS_01/01_Introduction}
\input{MS_01/02_RelatedWork}

\input{MS_01/03_MSDDP}

\input{MS_01/04_MHPC}

\input{MS_01/05_VWBC}

\input{MS_01/06_Implementation}
\input{MS_01/07_Result}

\input{MS_01/08_Conclusion}

\section*{Ackowledgements}
We would like to thank David Kelly, Shenggao Li, Nicolas Adrian, John Nganga, and Xuemin Liu for their help on the experimental setup and video shooting, Prof. Hai Lin and Zihao Song for lending the lab space, and Dr. Tingnan Zhang and Dr. Wenhao Yu for their helpful discussions and feedback on the manuscript.

The Mini Cheetah is sponsored by the MIT Biomimetic Robotics Lab and NAVER LABS.

\bibliographystyle{IEEEtran}
\bibliography{ms.bib}

\input{MS_01/Appendix_01}

\end{document}

%% file: preamble.tex
\usepackage{graphics,graphicx} 
\graphicspath{{../Figures/}}
\usepackage{amsmath} 
\usepackage{amssymb}  
\usepackage{xcolor}

\usepackage{amsmath} 
\usepackage{amssymb}  
\usepackage{amsthm}
\usepackage{xcolor}
\usepackage{algorithm}
\usepackage{algpseudocode}
\usepackage{cite}
\usepackage{dblfloatfix}
\usepackage{mathtools} 
\usepackage{dutchcal} 
\usepackage{caption}
\usepackage{subcaption}
\usepackage{hyperref}
\usepackage{lipsum}
\usepackage{array} 
\usepackage[bottom]{footmisc}

\usepackage{scalerel}
\newcommand{\norm}[1]{\left\lVert#1\right\lVert}
\newcommand{\snorm}[1]{\norm{#1}^2}

\newcommand{\x}{\mathbf{x}}

\newcommand{\f}{\mathbf{f}}

\newcommand{\fxx}{\mathbf{f}_{\mathbf{_{xx}}}}
\newcommand{\fuu}{\mathbf{f}_{\mathbf{{uu}}}}
\newcommand{\fux}{\mathbf{f}_{\mathbf{{ux}}}}

\newcommand{\inv}[1]{{#1}^{-1}}

\newcommand{\q}{\mathbf{q}}
\newcommand{\qd}{\Dot{\q}}
\newcommand{\qdd}{\Ddot{\q}}
\newcommand{\U}{\mathbf{U}}
\newcommand{\g}{\mathbf{g}}

\newcommand{\blambda}{\boldsymbol{\lambda}}

\newcommand{\s}{\mathbf{s}}

\newcommand{\p}{\mathbf{p}}
\newcommand{\T}{\mathbf{T}}
\newcommand{\Q}{\mathbf{Q}}
\newcommand{\Qu}{\Q_{\u}}
\newcommand{\Qx}{\Q_{\x}}
\newcommand{\Quu}{\Q_{\u\u}}
\newcommand{\Qux}{\Q_{\u\x}}
\newcommand{\Qxx}{\Q_{\x\x}}
\newcommand{\R}{\mathbf{R}}
\newcommand{\dx}{\delta\x}
\newcommand{\du}{\delta\u}

\newcommand{\X}{\mathbf{X}}
\newcommand{\A}{\mathbf{A}}
\newcommand{\B}{\mathbf{B}}

\newcommand{\W}{\mathbf{W}}
\newcommand{\K}{\mathbf{K}}
\newcommand{\J}{\mathbf{J}}

\newcommand{\taub}{\boldsymbol{\tau}}

\newcommand{\M}{\mathbf{M}}
\newcommand{\n}{\mathbf{n}}
\newcommand{\F}{\mathbf{F}}
\newcommand{\I}{\mathbf{I}}
\newcommand{\h}{\mathbf{h}}
\newcommand{\bomega}{\boldsymbol{\omega}}
\newcommand{\btheta}{\boldsymbol{\theta}}

\newcommand{\pib}{\boldsymbol{\pi}}

\newcommand{\grad}[2]{\frac{\partial {#1}}{\partial {#2}}}
\newcommand{\set}[1]{\mathbb{{#1}}}

\renewcommand{\S}{\mathbf{S}}

\renewcommand{\u}{\mathbf{u}}
\renewcommand{\P}{\mathbf{P}}
\renewcommand{\S}{\mathbf{S}}
\renewcommand{\u}{\mathbf{u}}
\renewcommand{\c}{\mathbf{c}}

\renewcommand{\d}{\mathbf{d}}
\renewcommand{\d}{\mathbf{d}}
\renewcommand{\r}{\mathbf{r}}
\newcommand{\part}{\partial}
\renewcommand{\xi}{\x^{[i]}}
\newcommand{\ui}{\u^{[i]}}

\newcommand{\trans}[1]{{#1}^{\top}}

\newcommand{\optdu}{\delta\Tilde{\u}}

\newcommand{\etal}{\textit{et al. }}
\newcommand{\eye}{\mathbf{1}}
\newcommand{\cmpc}{{\sc Cafe-Mpc }}
\newcommand{\dt}{d\text{t}}
\DeclareMathOperator*{\argmin}{arg\,min}

\newcommand{\as}{\mathcal{S}}
\newcommand\mat[1]{\begin{bmatrix}#1\end{bmatrix}} 

\newtheorem{remark}{Remark}

%% file: MS_01/00_Abstract.tex
\begin{abstract}
This work introduces an optimization-based locomotion control framework for on-the-fly synthesis of complex dynamic maneuvers. At the core of the proposed framework is a cascaded-fidelity model predictive controller ({\sc Cafe-Mpc}). {\sc Cafe-Mpc} strategically relaxes the planning problem along the prediction horizon (i.e., with descending model fidelity, increasingly coarse time steps, and relaxed constraints) for computational and performance gains.  This problem is numerically solved with an efficient customized multiple-shooting iLQR (MS-iLQR) solver that is tailored for hybrid systems. The action-value function from {\sc Cafe-Mpc} is then used as the basis for a new value-function-based whole-body control (VWBC) technique that avoids additional tuning for the WBC.
In this respect, the proposed framework unifies whole-body MPC and more conventional whole-body quadratic programming (QP), which have been treated as separate components in previous works. 
We study the effects of the cascaded relaxations in {\sc Cafe-Mpc} on the tracking performance and required computation time. We also show that the \cmpc , if configured appropriately, advances the performance of whole-body MPC without necessarily increasing computational cost. Further, we show the superior performance of the proposed VWBC over the Riccati feedback controller in terms of constraint handling. The proposed framework enables accomplishing for the first time gymnastic-style running barrel roll on the MIT Mini Cheetah. Video \href{https://youtu.be/YiNqrgj9mb8}{\underline{link}}.
\end{abstract}

\begin{IEEEkeywords}
Legged Locomotion, Model Predictive Control, Whole-Body Control, Differential Dynamic Programming
\end{IEEEkeywords}

%% file: MS_01/01_Introduction.tex
\section{Introduction}

\IEEEPARstart{U}{nlocking} biological-level mobility on legged robots is helpful to understand and discover their full potential for applications. Significant progress has been made over the past decades on both very capable hardware platforms and advanced control techniques. The MIT Cheetah-series robots demonstrated robust stair climbing skills \cite{bledt2018cheetah}, jumping over obstacles \cite{park2017high}, and back-flipping maneuvers\cite{katz2019mini}. The ETH AnyMal-series robots showed remarkable capabilities with stepping stones \cite{grandia2021multi, grandia2023perceptive}, and traversing extremely unstructured environments\cite{hwangbo2019learning, lee2020learning}. Researchers from KAIST enabled quadruped robots to walk on vertical walls \cite{hong2022agile} and deformable soft terrains \cite{choi2023learning}. The hydraulic-actuated HyQ-series robots were shown to drag a 3-ton airplane \cite{HyQReal}, and with terrain adaptation capability \cite{villarreal2020mpc}.

Despite the rapid progress, achieving biological levels of mobility on robots remains difficult. Well-trained human professionals can perform parkour and gymnastics that involve significant body rotations (barrel roll, flip, etc) in the middle of running. This level of mobility is a challenge for robots, and has not been shown on any quadruped platforms in the literature. The challenges are two fold. First,  gymnastic-style motions require careful coordination of the whole body. When using a model-based approach, the whole-body dynamics poses a challenge to satisfy real-time computation constraints, due to its non-convexity and high dimension. Second, the controller should be sufficiently flexible and robust to smoothly synthesize different motions (e.g., running, in-air body rotations, etc.), and the transitions between them.

Prior studies have made important progress toward attaining dynamic acrobatic maneuvers on quadruped robots. Offline trajectory optimization (TO) with a detailed dynamics model including the motors was used in \cite{katz2019mini} to achieve a back flip with the MIT Mini Cheetah. Due to its offline nature, the robot needed to be sufficiently close to a proper initial condition for successful execution. Tuning of the tracking controller was further treated as a fully separate component. Online TO with single-rigid body (SRB) models has been studied \cite{chignoli2021online, song2022optimal} to accomplish an in-place barrel roll. Since leg momentum is ignored, the SRB inertia needs to be tuned as a proxy for the leg inertia in order to obtain sufficient take-off velocity. In addition to the model-based method, Li \etal \cite{li2023learning} attempted to learn a back-flipping policy from partial demonstration. Regardless, the acrobatic maneuvers attained in all previous works require intermediate static starting and landing poses. 

\begin{figure}
    \centering
    \includegraphics[width=0.8\linewidth]{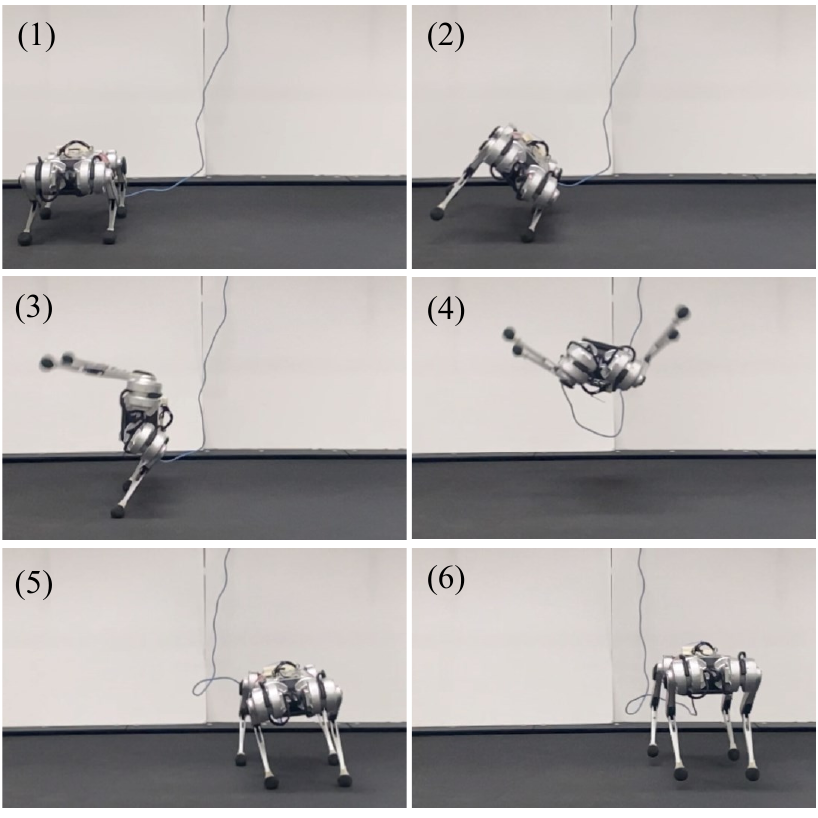}
    \caption{In-place barrel roll on the MIT Mini Cheetah accomplished with the proposed control framework. The robot performs an in-place barrel roll, followed by a hopping step, and a pacing gait. All motions and transitions are fully synthesized online. The results section includes more challenging tasks where the robot performs a barrel roll in the middle of running.}
    \label{fig:inplace-br-intro}
\end{figure}

This work aims to further push the mobility of quadruped robots. We propose an optimization-based control framework that takes as its input a reference trajectory obtained from a motion library, and then outputs commands that are directly executable on the robot hardware. Apart from diverse regular locomotion skills such as pacing and bounding, this framework unlocks on-the-fly synthesis of a gymnastic running barrel roll. An in-place barrel roll reference is used, and the controller is capable of adapting it online to several running gaits with different initial contact configurations. The proposed framework makes it easy to create diverse behaviors by simply specifying the references. Figure~\ref{fig:inplace-br-intro} highlights one of the main results achieved in this work.

\begin{figure*}[t]
    \centering
    \includegraphics[width=0.75\linewidth]{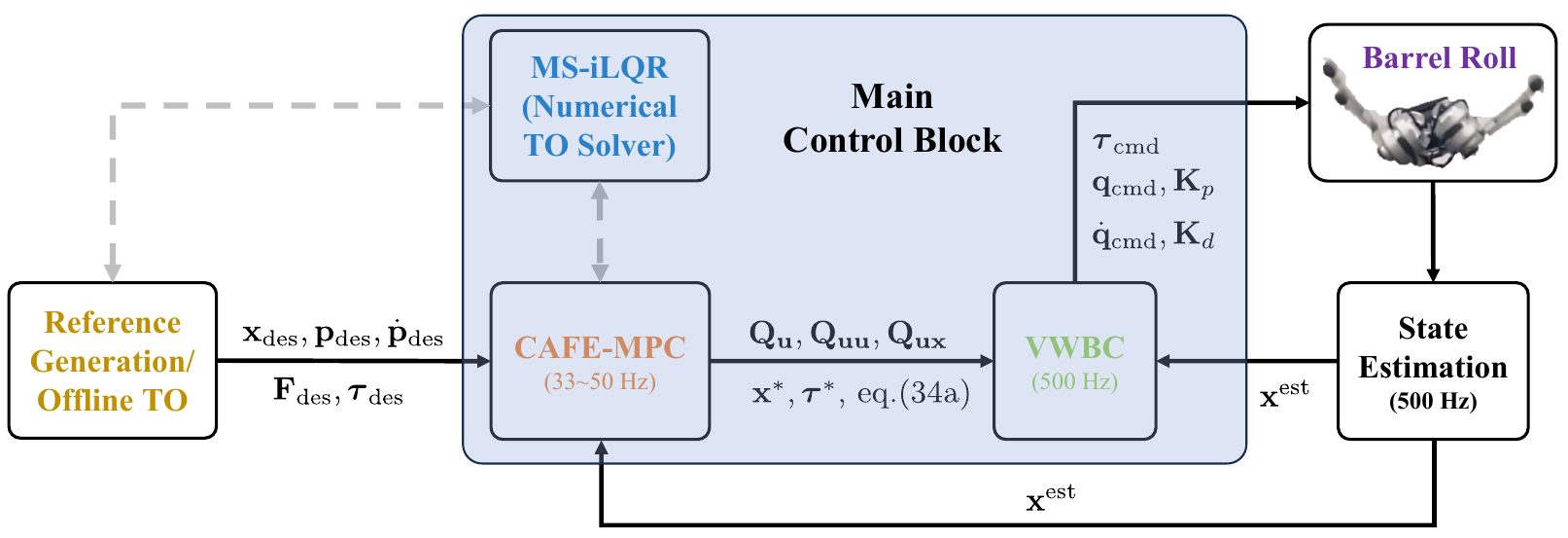}
    \caption{An overview of the system architecture. The proposed control framework takes a reference trajectory as input, and outputs commands that are directly executable on the robot. The motion compiler consists of the {\sc Cafe-Mpc}, the customized MS-iLQR solver for numerical optimization, and the VWBC. The MS-iLQR solver is used for offline TO as well. The MPC shares the same cost function for all tasks, and the VWBC is tuning-free.}
    \label{fig:overview}
\end{figure*}

The proposed framework is built upon model predictive control (MPC), a method of producing control inputs by predicting the future behavior of the robot as part of solving a trajectory optimization (TO) problem. MPC accounts for model uncertainties and external disturbances by frequently adapting the predicted plan to the most recent state. Whole-body MPC produces a high-fidelity plan that is coherent with the full robot dynamics. 
However, it is known to have large computational burdens, prohibiting the use of a long prediction horizon or fast update frequency. 
For this reason, the capability of this approach to control highly dynamic motions on robot hardware has not yet been demonstrated \cite{mastalli2023inverse, dantec2022whole, jallet2023proxddp}. A local feedback Riccati controller \cite{mastalli2023inverse} that runs at much higher frequency can be used to account for the policy delay and the slow MPC update. This feedback controller, however, is not guaranteed to respect the constraints in the whole-body MPC. A notable strategy to alleviate the computational burden is to instead use a simplified template model, thus template MPC. A low-level controller that converts the low-dimensional plan to whole-body commands is needed, often an inverse-dynamics quadratic program (QP). This strategy is thus far the most popular MPC approach, and has demonstrated great success for legged locomotion \cite{di2018dynamic, villarreal2020mpc, grandia2023perceptive, meduri2023biconmp, chen2023beyond, li2023dynamic, gu2023walking}. However, the omitted model details and the possibly added artificial constraints may result in infeasible or over-conservative target plans \cite{wensing2023optimization}. As a result, nontrivial tuning of the whole-body QP is often needed. 
The proposed framework introduces a novel approach that accelerates the whole-body MPC, and a novel low-level controller that integrates the Riccati controller and the whole-body QP.


\subsection{Contributions}

The contributions of this work are summarized as follows. The first contribution is a cascaded-fidelity model predictive controller ({\sc Cafe-Mpc}). 
{\sc Cafe-Mpc} generalizes our previous MHPC ideas beyond model cascades, exploring the use of increasingly coarse integration time steps and progressively relaxed constraints along the prediction horizon. The second contribution is to extend multiple-shooting iLQR \cite{li2023unified} to a class of hybrid systems whose phase sequence and timings are fixed. Specifically, we focus on how value function approximations can be back-propagated and how shooting nodes are updated across the switching surfaces. The third contribution is a value-function-based whole-body controller (VWBC). The VWBC employs a local action-value function as its minimization objective, ensuring a direct link between MPC and the WBC objectives. This makes it tuning-free, as opposed to the non-trivial additional cost design needed for conventional WBCs. The resulting scheme of {\sc Cafe-Mpc} + VWBC provides a structure that unifies whole-body MPC and whole-body QPs, which were conventionally implemented as separate designs. 

The last contribution is the overall optimization-based control framework. Beyond regular locomotion skills, this framework unlocks real-time synthesis of gymnastic-style motions, by simply specifying an input reference motion without further tuning of the system parameters. With the developed framework, we show a quadruped robot can achieve a barrel roll in the middle of running. To the best of our knowledge, this is the first time a running barrel roll has been accomplished on quadruped hardware. Figure~\ref{fig:inplace-br-intro} highlights the result of an in-place barrel roll on the MIT Mini Cheetah, while more challenging tasks are described in Section~\ref{sec:result_br}. We further plan to open-source our code with the publication of the manuscript, and hope it can be helpful for the readers and to the robotics community.

\subsection{System Overview and Outline}
An overview of the system architecture is shown in Fig.~\ref{fig:overview}. The overall control framework takes as input a reference trajectory, and outputs the whole-body commands (joint torques, angles, and velocities) which are directly executable on the robot. For regular locomotion skills like trotting and bounding, simple heuristic reference trajectories are sufficient. For combined motions with the barrel roll, a long-horizon TO is solved offline to provide a more detailed motion sketch that can then be generalized to new situations. For example, the in-place barrel roll reference can be combined with a pacing reference, with the online synthesis of a running barrel roll then left to our framework. The {\sc Cafe-Mpc} runs at 33-50 Hz, and the VWBC refines the MPC command at 500 Hz. For both offline TO and online MPC, the customized MS-iLQR is employed as the underlying numerical solver.

The rest of the paper is structured as follows. In Section~\ref{sec:related_work}, we summarize related works that are associated with the core components of Fig.~\ref{fig:overview}. Section~\ref{sec:MSDDP} first reviews our previous work on MS-iLQR, and discusses its extension to constrained multi-phase optimal control problems. In Section~\ref{sec:MHPC}, we present the detailed formulation of the {\sc Cafe-Mpc}. Section~\ref{sec:VWBC} proposes the novel VWBC, and discusses its relations to conventional controllers. In Section~\ref{sec:implementation}, we present the heuristic reference generation, offline TO for the barrel roll, and other details used to complete the blocks in Fig.~\ref{fig:overview}. Section~\ref{sec:results} discusses the simulation and hardware results. Section~\ref{sec:conclusion} discusses limitations, and concludes the paper with suggestions for future work.

%% file: MS_01/02_RelatedWork.tex
\section{Related Work}
\label{sec:related_work}

\subsection{MPC for Legged Robots}

Model Predictive Control (MPC) provides a means of controlling a robotic system by predicting its future behaviors. The expected behaviors (e.g., maintaining balance) are achieved via minimizing an objective function subject to some modeled dynamics and constraints, which are often nonlinear. 
The capability of MPC to cope with nonlinear dynamics and constraints makes it well-suited and increasingly popular for the control of legged robots, as evidenced by a growing body of literature \cite{tassa2012synthesis,wieber2006trajectory,di2018dynamic, neunert2018whole, bledt2019implementing, kuindersma2016optimization, apgar2018fast, villarreal2020mpc}.

The classical template-MPC approach generates plans of low dimensions. Therefore, a low-level controller is needed to (1) produce the whole-body commands (2) provide fast feedback control for stabilization. One such notable controller is an inverse-dynamics quadratic program (QP) \cite{hutter2014quadrupedal}. Popular template models include the Linear Inverted Pendulum Model (LIP) \cite{wieber2006trajectory, gao2023time}, the Spring-Loaded Inverted Pendulum Model (SLIP) \cite{wensing2013high}, Single-Rigid-Body Model \cite{bledt2017policy,winkler2018gait,di2018dynamic}, and the centroidal model \cite{orin2013centroidal,dai2014whole, wensing2016improved, romualdi2022online, grandia2023perceptive}. Template MPC has the advantage of fast computation due to the relatively small and possibly convex optimization problems. 
One limitation of this approach, however, is that the operational envelope of the resulting motions either underestimates or overestimates the set of whole-body feasible motions. As a result, the resulting target plan may not be tractable by the low-level whole-body controller. A recent survey paper provides a more detailed overview of this perspective \cite{wensing2023optimization}.


The whole-body MPC approach does not have the problem of producing infeasible trajectories of the template MPC. This approach, however, is notoriously known for its computational burden, due to the increased dimensions, non-linearity, and non-convexity. For these reasons, numerous efforts have been made to accelerate the whole-body MPC by developing efficient structure-exploiting solvers \cite{neunert2018whole, dantec2022whole, mastalli2022feasibility, katayama2023structure, jordana2023stagewise}, fast analytical dynamics \cite{carpentier2018analytical, singh2022efficient}, and by using inverse dynamics \cite{mastalli2023inverse}. Several works have shown the success of this approach on robot hardware with regular locomotion skills \cite{dantec2022whole, mastalli2023inverse, jallet2023proxddp}. However, highly dynamic behaviors have yet been demonstrated on any hardware platform.

Our prior work on MHPC \cite{li2021model} combines the benefits of whole-body MPC and template-MPC, by placing a whole-body model in the near term and a template model in the distant term. This similar idea was further explored in several follow-up works \cite{wang2021multi, liu2022design, khazoom2022humanoid} with different models that were tailored towards humanoid robots. While the effect of model schedule on disturbance rejection is studied in \cite{li2021model}, Kahzoom \etal \cite{khazoom2023optimal} proposed a method to optimize the model schedule. Following a related but distinct idea, Norby \etal \cite{norby2022adaptive} fixed the prediction horizon while adapting the model fidelity based on the task complexity. The \cmpc proposed in this work further relaxes the template plan using coarse timesteps and relaxed constraints. The motivation is that the template model has slower dynamics (e.g., due to ignoring fast swinging appendages), and the constraints in the distant end are less critical to the current decisions. Similar ideas have also been studied beyond the legged robot community, such as for chemical process control \cite{kameswaran2012multi, shin2021diffusing}, with simple mobile robots \cite{bathge2016exploiting,brudigam2021model}, and with autonomous vehicles \cite{Laurense2022}.

Contact-implicit MPC is another line of compelling research for legged locomotion, which enables simultaneous generation of the contact modes and the whole-body motions. Prior studies that leverage differentiable contact models have shown the promise of this approach of generating complex multi-contact behaviors \cite{tassa2012synthesis, kim2023contact, kurtz2023inverse, le2024fast}. This approach, however, is known to have numerical problems such as the difficulty to obtain good initial guesses, and computational burden. Our work is complementary to the contact-implicit approach, in the sense that the contact-implicit planning could be considered as a low-frequency top layer that informs the \cmpc with contact sequences, and leverages \cmpc for fast online synthesis.

\subsection{Numerical Optimization for MPC}
One critical factor for the success of MPC is to reliably and efficiently solve the underlying TO problems. Conventional numerical optimization methods for TO take one of the three approaches \cite{diehl2006fast}: \textit{dynamic programming}, \textit{indirect methods}, and \textit{direct methods}.  
\textit{Direct methods} are the most widespread in robotics, and they proceed by transcribing the TO problems to a Nonlinear Programming (NLP) problem which could be effectively solved using well-developed off-the-shelf NLP solvers, for instance, SNOPT \cite{gill2005snopt} or IPOPT \cite{wachter2006implementation}. Most NLP solvers proceed by taking successive linearization of the KKT condition, resulting in complexity of $O(N^3)$ (in the worst case) where $N$ is the problem size. Consequently, this approach is prohibitive for online use with robots where $N$ is large. Fortunately, some \textit{direct methods} such as multiple shooting and direct collocation result in sparse NLPs whose computation complexity could be reduced to $O(N)$ \cite{betts2010practical}. For this reason and thanks to the development of computation hardware, \textit{direct methods} have been employed by some research groups to successfully solve nonlinear MPC for legged robots \cite{kuindersma2016optimization, bledt2019implementing, grandia2023perceptive, jordana2023stagewise}. It was shown in \cite{grandia2023perceptive} that the nonlinear MPC can run up to 100 Hz with a real-time iteration scheme \cite{diehl2005real} and an interior-point-method (IPM) based NLP solver \cite{frison2020hpipm}.


Differential Dynamic Programming (DDP) \cite{mayne1966second} (also known as iLQR when adopting a Gauss-Newton Hessian approximation \cite{tassa2012synthesis}) is a powerful tool for nonlinear optimal control and has gained increasing attention in the past decade in the robotics community. It successively solves a sequence of small sub-problems leveraging Bellman's equation. Similar to the structure-exploiting sparse NLP solvers \cite{frison2020hpipm, katayama2023structure, jordana2023stagewise}, DDP has a linear computational cost relative to the prediction horizon. However, it comes with a value function approximation and a local feedback policy for free as intermediate results, which can be used for higher-rate lower-level control \cite{grandia2019feedback, mastalli2023inverse}. These properties make DDP well-suited for MPC of legged robots. 

Many research groups have made efforts toward this direction. Tassa \etal achieved complex motion control with iLQR on a simulated humanoid robot, with slow simulation speed \cite{tassa2012synthesis}. Koenemann \etal implemented the same framework on the humanoid robot HRP-2 \cite{koenemann2015whole}. Though the task was simple (balance standing), it was the first time whole-body MPC was achieved on humanoid robot hardware. Neunert \etal accomplished locomotion control on two high-performance quadruped robots, ANYmal and HyQ, in simulation \cite{neunert2017trajectory} and on hardware \cite{neunert2018whole}, where iLQR served as the underlying solver. Though promising, there are several aspects where opportunities for improvement are clearly seen. 
Effective constraint handling and sensitivity to initial guesses have been the major bottlenecks. A large amount of recent works thus have contributed to addressing these problems. To deal with constraints, DDP/iLQR is combined with many common algorithms in the numerical optimization community, for instance, Projected Newton method \cite{tassa2014control}, penalty method \cite{farshidian2017efficient}, barrier method \cite{grandia2019feedback} and interior-point method\cite{pavlov2021interior}, Augmented Lagrangian (AL) method (e.g., \cite{Howell19, lantoine2012hybrid, plancher2017constrained}) and primal-dual AL \cite{jallet2023proxddp}. To deal with the sensitivity problem, DDP /iLQR is extended with multiple-shooting formulations that permit an infeasible warm start with a state trajectory. \cite{giftthaler2018family, Howell19, mastalli2022feasibility, li2023unified, jallet2023proxddp}. 

\subsection{Whole-Body Control}
The whole-body control (WBC) technique is an important component in many MPC-based control frameworks. It operates at a higher loop frequency, and produces commands that are directly executable on the robot hardware. Multiple WBC techniques have been developed in the literature. 
The common approach is to formulate the WBC as a quadratic programming (QP) problem. The objective function of the QP consists of a list of weighted tasks specified in the operational space \cite{sentis2005synthesis, hutter2014quadrupedal, kuindersma2014efficiently, apgar2018fast, grandia2023perceptive}, thus OSC-QP. These tasks are designed to track certain references in the acceleration level in the least-square sense, such as CoM and swing foot accelerations, etc. It was shown in \cite{kim2019highly} that the dynamic capability of a quadruped can be significantly improved by incorporating the GRF in the objective. The OSC-QP reasons about the whole-body dynamics, and enforces constraints such as torque limits, non-slipping constraints, etc. As such, it offers a more feasible control. 
The OSC-QP requires non-trivial tuning, which arises from three aspects. First, the relative importance between tasks is based on heuristics. Second, the tasks are often designed using simplified models, which are inconsistent with the whole-body dynamics in OSC-QP. Third, the OCS-QP is designed for a short term (one control step). The long-term stability of the OSC-QP relies on the quality of the reference, which is regulated by a PD controller, and requires further tuning.

The use of a Riccati controller is a less-known but powerful WBC technique. The feedback gains can be computed either via DDP/iLQR \cite{grandia2019feedback, li2022versatile, mastalli2023inverse}, or via multiple-shooting based numerical optimization \cite{grandia2023perceptive}, though the ideas are similar where the Difference Riccati equations are computed. The Riccati controller can either be directly used if the MPC reasons about whole-body dynamics \cite{mastalli2023inverse}, or an additional layer is needed to produce the whole-body actuation if the MPC reasons about simplified models \cite{grandia2019feedback, li2022versatile}. A known problem of this approach is that the feedback control does not necessarily satisfy constraints, which can interfere with robot stability, for instance, when the friction cone constraints are violated.

The value-based WBC (VWBC) proposed in this work overcomes the issues of the conventional OSC-QP and the Riccati controller. The VWBC is not significantly different from the OSC-QP in terms of formulation. The novelty is that it employs an objective function that considers the long-term cost-to-go (value function). This value function is a proxy of the long-term stability, and is readily available from the top-level {\sc Cafe-Mpc}, thus leaving the VWBC free of additional tuning. The VWBC generalizes the Riccati controller to include constraints, thereby unifying whole-body MPC with conventional WBC. 

The most relevant works in the literature to VWBC are \cite{kuindersma2014efficiently, chignoli2020variational}. The value functions in these works are approximated based on simplified models, which reflect the long-term balance only in a low-dimensional state space. Consequently, additional tracking terms and regularization terms are needed to produce whole-body coordination. The value function used by the VWBC in this work, however, reasons about the whole-body model, which does not require additional cost terms.

\subsection{RL for Legged Robots}
Reinforcement Learning (RL) is yet another powerful technique for the control of legged robots. Instead of performing heavy online optimization as in MPC, RL trains a control policy offline in simulated environments. The control policy can be made very robust (either in the sense of sim-to-real transfer or uncertain environments) by injecting noise during training into the robot dynamics, the environments, and the sensory information (encode measurements, depth images, etc). As such, great success has been shown with RL that enables quadruped robots to traverse extremely unstructured environments \cite{hwangbo2019learning, lee2020learning, miki2022learning}, imitate animal locomotion behaviors \cite{peng2020learning}, run on slippery ground \cite{ji2022concurrent}, and walk over deformable terrains \cite{choi2023learning}.

Despite the remarkable progress of RL on robust quadruped locomotion, it remains challenging to achieve highly dynamic animal-like behaviors on hardware. Many researchers thus have shifted gears towards agile quadruped locomotion. Recent works \cite{hoeller2023anymal, zhuang2023robot, cheng2023extreme} accomplished parkour-like maneuvers with RL on quadruped robots. Though impressive versatility and agility are achieved, the gap remains in terms of dynamic capabilities when compared to the jumping-over-obstacle behaviors previously achieved with the optimization-based approach \cite{park2017high}, not to mention the biological-level mobility. Other works show fast locomotion \cite{margolis2022rapid} and continuously dynamic jumping \cite{yang2023cajun}, which focus on regular locomotion gaits, while versatility and agility remain to be explored. 


A recent work \cite{jenelten2024dtc} shows the promise of advancing RL using MPC. Though a pure RL policy enables robust locomotion, precision (such as foot placements) has been a problem. It was shown in \cite{jenelten2024dtc} that both robustness and precision can be achieved by guiding RL with MPC-planned foot placements. Since our work advances MPC, it can potentially be used to advance learning-based frameworks such as \cite{jenelten2024dtc} as well.

%% file: MS_01/03_MSDDP.tex
\section{Multiple-Shooting Differential Dynamic Programming}
\label{sec:MSDDP}

The {\sc Cafe-Mpc} considers multiple dynamics phases along the planning horizon due to the cascaded model fidelity and the change of contact status, resulting in a hybrid system. We assume this hybrid system has a fixed phase (dynamics) sequence and timings. Optimal control over such hybrid systems can be modeled by a multi-phase TO problem \cite{lantoine2012hybrid}.  In this section, we present an efficient customized multiple-shooting DDP (MS-DDP) solver for generic multi-phase TO problems. We then discuss how to construct a multi-phase TO problem from {\sc Cafe-Mpc} in Section~\ref{sec:MHPC}.

Let $n_p$ denote the number of phases, and $i$ denote the phase index. We model such systems in discrete time as below \cite{li2020hybrid}
\begin{equation}\label{eq:switchDyn}
    \begin{cases}
     \xi_{k+1} &= \f_i(\xi_k,\ui_k)\\ 
     \x^{[i+1]}_0 &= \P_{i}(\xi_{N_{i}})
    \end{cases},
\end{equation}
where $\x$ denotes the state variable, $\u$ the control variable, $\f_i(\cdot,\cdot)$ the phase dynamics, and $\P_i(\cdot)$ is the reset map from the current phase to the next phase, $N_i$ denotes the number of time steps of the $i^{\text{th}}$ phase. Note that the dimensions of $\x$ and $\u$  may vary between phases. A constraint is applied at the end of each phase
\begin{equation}\label{eq:terminalConstr}
    \g_i(\xi_{N_i}) = 0.
\end{equation}
The switching constraint \eqref{eq:terminalConstr} (also known as phase terminal constraint \cite{lantoine2012hybrid}) encodes state-based switching. It requires that the trigger of the reset map is conditioned on satisfying the constraint~\eqref{eq:terminalConstr}. For legged robots, for instance, the impact dynamics is triggered only when a foot touches the ground (foot height is zero). Without Eq.~\eqref{eq:terminalConstr}, the system~\eqref{eq:switchDyn} represents a time-based switched system. Details can be found in \cite{li2020hybrid}.

Giftthaler \etal\cite{giftthaler2018family} and Mastali \etal\cite{mastalli2020crocoddyl} introduced two seminal implementations of MS-DDP. Our previous work \cite{li2023unified} provides a framework that unifies these previous methods and offers a few more advancements. Nevertheless, these past implementations were either presented for single-phase systems \cite{giftthaler2018family}, or were not designed to cope with state-based switching constraints. In this section, we advance the MS-DDP for TO of hybrid systems (with fixed phase sequence and time), specifically dealing with the reset maps and the constraints. We first review the background on MS-DDP for single-phase TO. We then discuss how to incorporate the reset maps in MS-DDP for unconstrained multi-phase TO, and finally introduce methods to deal with the constraints.

\subsection{MS-DDP for Single-Phase TO}\label{subsec:single-phase}
A single-phase TO problem has the form
\begin{subequations}\label{eq:TO}
\begin{IEEEeqnarray}{rl}
\min_{\U, \X} \ \ & J(\X, \U) = \sum_{k=0}^{N-1} \ell_k(\x_k, \u_k) + \phi(\x_N)\label{eq:TO_cost} \\
\text{subject~to} \ \ & \f(\x_k, \u_k)-\x_{k+1}  =  0 \label{eq:TO_dyn}
\end{IEEEeqnarray}
\end{subequations}
where $k$ ($0 \leq k \leq N$) denotes the time index with $N$ the prediction horizon length, $\u_k \in \set{R}^m$ is the control variable, $\x_k \in\set{R}^n$ denotes the state variable, $\U = \{\u_k\}_{k=0}^{N-1}$ and $\X = \{\x_k\}_{k=0}^N$ respectively, stack the controls and states along the trajectory, $\ell_k(\x_k, \u_k)$ and $\phi(\x_N)$ respectively are the running cost and the terminal cost, and $\f(\x_k, \u_k)$ is the system dynamics. In Eq.~\eqref{eq:TO}, we omit the dependency on the phase index $i$ for clarity as it is a single-phase problem.

In the multiple-shooting setting, both $\X$ and $\U$ are decision variables, whereas in the single-shooting setting, $\U$ is the decision variable, and $\X$ depends on $\U$ via system dynamics. Denote $\Bar{\X}$, $\Bar{\U}$ the current estimates for the optimal solution (also known as the nominal values) of $\X$ and $\U$. The simulated state $\f(\Bar{\x}_k, \Bar{\u}_k)$ and the estimate $\Bar{\x}_{k+1}$ are likely different. This difference is also known as the defect \cite{giftthaler2018family, mastalli2022feasibility}, and is defined as
\begin{equation}
    \Bar{\d}_{k+1} := \f(\Bar{\x}_k, \Bar{\u}_k)-\Bar{\x}_{k+1}.
\end{equation}
Giftthaler\cite{giftthaler2018family} proposes a flexible formulation that allows a subset of $\X$ to be independent while keeping the rest dependent variables. These independent variables are called shooting states, the indices of which are collected to a set $\set{M}$. These dependent variables are called roll-out states, the indices of which are collected to the complementary set $\Tilde{\set{M}}$. If $k\in \set{M}$, then $\Bar{\x}_k$ is seeded from an initial guess. Otherwise, it is overwritten by the simulated state $\f(\Bar{\x}_{k-1}, \Bar{\u}_{k-1})$. MS-DDP iteratively improves $(\Bar{\X}, \Bar{\U})$ until the defects are sufficiently small and the cost function is minimized. At each iteration, it performs a backward sweep followed by a forward sweep within a line search process.

\subsubsection{Backward Sweep} 
Leveraging Bellman's principle of optimality for discrete systems, the backward sweep performs a one-step optimization at every time step along the nominal trajectory, which produces a local optimal control policy. Denote $(\dx, \du)$ small perturbations to $(\Bar{\x}, \Bar{\u})$, and $\pib$ the local optimal control policy. The one-step optimization is formulated as 
\begin{equation}\label{eq:delta_bellman}
    \pib_k(\Bar{\x}_k + \dx_k) = \argmin_{\du_k} \big(\underbrace{\delta\ell_k(\dx_k, \du_k) + v_{k+1}(\dx_{k+1})}_{Q_k(\dx_k,\du_k)} \big)
\end{equation}
where $\delta\ell_k(\cdot,\cdot)$ is the variation of $\ell_k(\cdot,\cdot)$ in the neighborhood of $(\Bar{\x}_k, \Bar{\u}_k)$ due to $(\dx_k, \du_k)$, $v_{k+1}$ is the local value function (optimal cost-to-go) approximation for the perturbed state at time $k+1$, and $Q_k(\cdot, \cdot)$ denotes the local action-value function at time $k$. The function $Q_k(\cdot, \cdot)$ cannot generally be represented in closed form, and thus in practice is approximated to the second order. Temporarily omitting the subscript $k$, we have
\begin{multline}\label{eq:Q_approx}
        Q(\dx, \du) \approx \\
        \frac{1}{2}\mat{\dx\\\du}^{\top}
        \mat{\Qxx & \trans\Qux\\\Qux & \Quu}
        \mat{\dx\\\du} + \trans{\Qx}\dx + \trans{\Qu}\du
\end{multline}
where $\Quu$ and $\Qux$ are second-order partials, and $\Qx$ and $\Qu$ are first-order partials. Performing the minimization in Eq.~\eqref{eq:delta_bellman} with Eq.~\eqref{eq:Q_approx} results in the locally optimal control policy
\begin{equation}\label{eq:control_policy}
    \pib_k(\x_k) = \underbrace{-\Q_{\u\u,k}^{-1}\Q_{\u,k}}_{\optdu_k} \underbrace{-\Q_{\u\u,k}^{-1}\Q_{\u\x,k}}_{\K_k}(\x_k - \Bar{\x}_k),
\end{equation}
where $\optdu$ and $\K$ are the optimal increment of the feed-forward control and the locally optimal feedback gain, respectively. At convergence, $\optdu\approx 0$ and the resulting policy is the Riccati feedback controller \cite{mastalli2023inverse}.
Let $\A = \left.\frac{\part \f}{\part \x}\right|_{(\Bar{\x},\Bar{\u})}$, $\B = \left. \frac{\part \f}{\part \u}\right|_{(\Bar{\x}, \Bar{\u})}$. Denote $\fxx, \fuu, \fux$ the tensors that represent the second-order partials of $\f$. Denote $\q_k$ and $\r_k$ the gradients of $\ell_k$ w.r.t. $\x$ and $\u$ respectively, $\Q_k$, $\R_k$, and $\P_k$ the second-order partials of $\ell_k$. The partials of $Q_k(\cdot, \cdot)$ along the trajectory are then calculated recursively using
\begin{subequations}\label{eq:Q_updates}
\begin{align}
    \hat{\s}_{k+1} &= \s_{k+1} +\S_{k+1}\Bar{\d}_{k+1} \\
    \Q_{\x,k} &= \q_k + \A_k^{\top}\hat{\s}_{k+1} \label{eq:Qs_DDP_Qx}\\
    \Q_{\u,k} &= \r_k + \B^{\top}_k\hat{\s}_{k+1} \label{eq:Qs_DDP_Qu}\\
    \Q_{\x\x,k} &= \Q_k + \A^{\top}_k\S_{k+1}\A_k + \s_{k+1}\cdot\fxx{}_{,k}  \label{eq:Qs_DDP_Qxx}\\
    \Q_{\u\u,k} &= \R_k + \B^{\top}_k\S_{k+1}\B_k + \s_{k+1}\cdot\fuu{}_{,k} \label{eq:Qs_DDP_Quu} \\
    \Q_{\u\x,k} &= \P_k + \B^{\top}_k\S_{k+1}\A_k  + \s_{k+1}\cdot\fux{}_{,k}\label{eq:Qs_DDP_Qux}
\end{align}
\end{subequations}
where $\S_k$, $\s_{k}$, and $s_k$ are the Hessian, gradient, and drift terms that quadratically approximate $v_k$ (Eq.~\eqref{eq:delta_bellman}) at the shooting state, and $\hat{\s}_k$ is an intermediate variable, the operator $\cdot$ denotes the vector-tensor multiplication. The value function approximations are recursively calculated using
\begin{subequations}\label{eq:V_updates}
\begin{align}
\S_k &= \Q_{\x\x,k} - \Q_{\u\x,k}^T \Q_{\u\u,k}^{-1} \Q_{\u\x,k} \label{eq_Vxx}\\
    \s_k &= \Q_{\x,k} - \Q_{\u\x,k}^T\Q_{\u\u,k}^{-1}\Q_{\u,k} \label{eq_Vx}\\
    s_k &= s_{k+1}  - \frac{1}{2}\Q_{\u,k}^T \Q_{\u\u,k}^{-1} \Q_{\u,k} 
    \label{eq_deltaV}   
\end{align}
\end{subequations} 
where the boundary conditions for Eqs.~\eqref{eq:Q_updates} and~\eqref{eq:V_updates} are
\begin{equation}\label{eq:V_boundary}
    \S_N = \Q_N, \s_N = \q_N, s_N = 0.
\end{equation} 
The Eqs.~\eqref{eq:Q_updates} and \eqref{eq:V_updates} become standard DDP backward sweep equations \cite{mayne1966second, tassa2012synthesis} when the defect is identically zero, and become multiple-shooting iLQR \cite{giftthaler2018family, mastalli2022feasibility} when the last terms of Eqs.~\eqref{eq:Qs_DDP_Qxx}-\eqref{eq:Qs_DDP_Qux} are removed. Our previous work \cite{li2023unified} enables unifying the backward sweep equations of all previous formulations \cite{mayne1966second, tassa2012synthesis, giftthaler2018family, mastalli2022feasibility}.

\begin{remark}
   The action-value function \eqref{eq:Q_approx} measures the effect of an action on the long-term cost-to-go given the current state. In the context of legged robots, the use of MPC was argued \cite{wieber2008viability} to imbue viability as a side effect when the cost function is properly designed. When MS-DDP is used to solve MPC, an optimal action-value function \eqref{eq:Q_approx} naturally results from the solution process, which allows local adjustment of the optimal open-loop control. With this motivation, Kuindersma \etal \cite{kuindersma2014efficiently} incorporated the optimal cost-to-go obtained from LQR for a simple model into a whole-body QP. In our case, the function \eqref{eq:Q_approx} acts on the whole-body states, providing a direct synergy between the \cmpc and the ultimate lower-level VWBC in Section~\ref{sec:VWBC}.
\end{remark}

\subsubsection{Forward Sweep} MS-DDP employs a hybrid forward roll-out \cite{giftthaler2018family} to update the nominal trajectory $(\Bar{\X}, \Bar{\U})$ by applying the control policy
\begin{equation}
    \u'_k = \Bar{\u}_k + \alpha\optdu_k + \K_k(\x'_k - \Bar{\x}_k).
\end{equation}
This control policy is first used to update $\Bar{\X}$ for $0 \leq k \leq N-1$ via linearized dynamics
\begin{equation}
     \x'_{k+1} = \Bar{\x}_k + \A_k(\x'_k - \Bar{\x}_k) + \B_k\du_k(\alpha) + \Bar{\d}_{k+1}. 
\end{equation}
where the superscript $'$ indicate variables after a trial step of size  $\alpha \in (0, 1]$, which is determined by a backtracking line search. Then the following equation is executed to sequentially overwrite the roll-out states, i.e., for $k+1\in \Tilde{\set{M}}$
\begin{equation}\label{eq:nonlinear_ls}
    \x'_{k+1} = \f(\x'_{k}, \u'_k)
\end{equation}
The defect variable and the cost function are then evaluated along the trial trajectory $(\X', \U')$. In practice, Eq.~\eqref{eq:nonlinear_ls} is evaluated in parallel, i.e., we simulate the dynamics starting from a shooting node until the next subsequent shooting node. 

\subsubsection{Line Search and Regularization}
\label{subsubsec:to-ls}
An adaptive merit function that automatically balances the defect and cost is used for backtracking line search. An exact expected cost change in the sense of linearized dynamics and quadratic cost approximation with the Armijo condition is used for the acceptance condition. Details on this regard are found in \cite{li2023unified}. A similar regularization method as in \cite{tassa2012synthesis} is used to ensure the positive definiteness of $\Quu$ in Eq.~\eqref{eq:Qs_DDP_Quu}. The result of a successful line search is that the nominal trajectory $\bar{\X}$, $\bar{\U}$ is updated, with the backward and forward sweeps then repeated to convergence.

\subsection{MS-DDP for Multi-Phase TO}
\label{subsec:uc-multi-phase}

\begin{figure}[t]
    \centering
    \includegraphics[width = 0.8\linewidth]{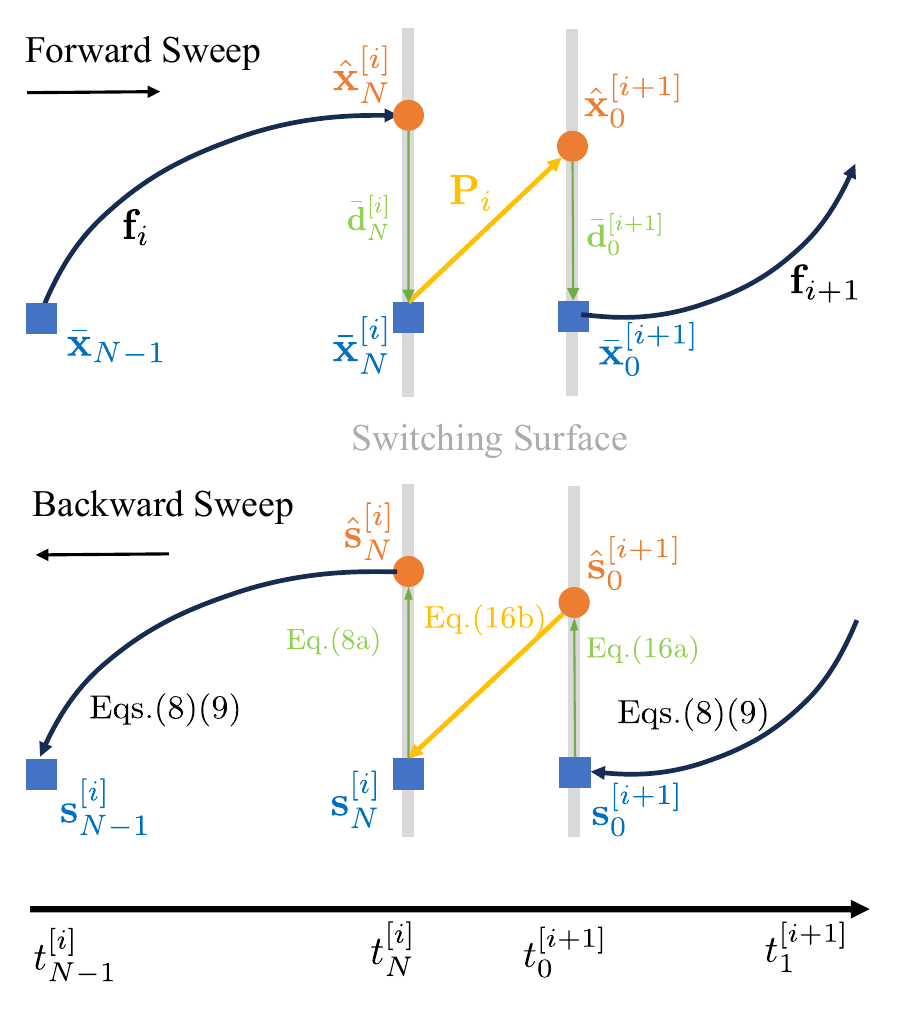}
    \caption{Illustration of forward sweep and backward sweep of MS-DDP for hybrid systems TO.}
    \label{fig:msddp_hybrid}    
\end{figure}

This section discusses the extension of MS-DDP to account for the hybrid system effects \eqref{eq:switchDyn} and \eqref{eq:terminalConstr}, which represents a contribution compared to our previous work \cite{li2023unified}. We first discuss the case without considering the switching constraint.

\subsubsection{Unconstrained Multi-Phase TO}
An unconstrained multi-phase TO problem is formulated as
\begin{subequations}\label{eq:u-multi-phase}
\begin{IEEEeqnarray}{cll}
\IEEEeqnarraymulticol{2}{l}{\min_{\X^{[\cdot]}, \U^{[\cdot]}} \ \ \sum_{i=1}^{n_p} J^{[i]}(\X^{[i]},\U^{[i]})}\label{eq:hybrid_cost}\\
    \text{subject~to} \ \ & \eqref{eq:switchDyn} 
\end{IEEEeqnarray}
\end{subequations}
where the variables with the superscript $^{[i]}$ or subscript $i$ indicate they are phase-dependent. The remaining variables share the same definitions as in problem~\eqref{eq:TO} and Eq.~\eqref{eq:switchDyn}.

The reset map instantaneously changes the state, potentially to a lower-dimensional space, involving a discontinuous (in time) jump. While the forward sweep and the backward sweep remain the same as in a single-phase problem until the end of a phase, care must be taken when performing forward sweep and backward sweep across the reset map. We define the defect after the reset map as
\begin{equation}
    \Bar{\d}^{[i+1]}_{0} = \P_i(\Bar{\x}^{[i]}_{N_i}) - \Bar{\x}^{[i+1]}_0
\end{equation}
which is evaluated in the forward sweep once the state trajectory $\X^{[i]}$ is initialized for each phase. In the backward sweep, we update the value functions across the reset map using
\begin{subequations}\label{eq:imp_bs}
    \begin{align}
        \hat{\s}^{[i+1]}_0 &= \s^{[i+1]}_0 + \S^{[i+1]}_0\Bar{\d}^{[i+1]}_0 \\         
        \s^{[i]}_N &= \q^{[i]}_N + \grad{\trans{\P}_i}{\x}\hat{\s}^{[i+1]}_0 \\
        \S^{[i]}_N &= \Q^{[i]}_N + \grad{\trans{\P}_i}{\x}\S^{[i+1]}_0 \grad{\P_i}{\x} + \s^{[i+1]}_0 \cdot \P_{\x\x,i}\label{eq:S_shooting}
    \end{align}
\end{subequations}
where $\P_{\x\x,i}$ is a tensor representing the second-order partials of $\P$. Same as Eq.~\eqref{eq:Q_updates}, omitting the last term in Eq.~\eqref{eq:S_shooting} results in iLQR/MS-iLQR. The Eq.~\eqref{eq:imp_bs} is an impact-aware backward step. It is analogous to HS-DDP in our previous work \cite{li2020hybrid}, but is extended to incorporate defects due to an infeasible warm start. Figure~\ref{fig:msddp_hybrid} graphically illustrates a forward sweep and a backward sweep across two consecutive phases.

\subsubsection{Constrained Multi-Phase TO}
We now consider the constrained multi-phase TO problems, which are formulated as
\begin{subequations}\label{eq:c-multi-phase}
\begin{IEEEeqnarray}{cll}
\IEEEeqnarraymulticol{2}{l}{\min_{\X^{[\cdot]},\U^{[\cdot]})} \ \ \sum_{i=1}^{n_p} J^{[i]}(\X^{[i]},\U^{[i]})}\label{eq:hybrid_cost}\\
    \text{subject~to} \ \ & \eqref{eq:switchDyn}, \eqref{eq:terminalConstr}\\
     &\h_i(\xi_k,\ui_k) \geq \mathbf{0}, \label{eq:g_ineq}
\end{IEEEeqnarray}
\end{subequations}
where $\h_i(\cdot):\set{R}^n\times\set{R}^m\rightarrow\set{R}^{n_h}$ represents general inequality constraints.  

We follow the previous work \cite{li2020hybrid} to take a bi-level approach to solve the problem~\eqref{eq:c-multi-phase} with MS-DDP. The terminal constraint~\eqref{eq:terminalConstr} is handled with Augmented Lagrangian (AL) \cite{nocedal2006numerical}, and the inequality constraint~\eqref{eq:g_ineq} is dealt with using the Relaxed Barrier (ReB) method \cite{hauser2006barrier}. Though similar in spirit to the previous work \cite{li2020hybrid}, the inner-loop optimization here is based upon the multiple-shooting formulation. We can now provide a reasonable initial guess of $\X$ so that the violation of the terminal constraint is not as bad as in the previous single-shooting formulation, facilitating faster convergence. Note that we omit the second-order terms in eqs.~\eqref{eq:Q_updates} and \eqref{eq:imp_bs} for real-time implementation, thus resulting in MS-iLQR, which we use in the rest of the paper. Work on incorporating second-order analytical dynamics \cite{singh2023second} is in progress.

%% file: MS_01/04_MHPC.tex
\section{Cascaded-Fidelity Model Predictive Control}
\label{sec:MHPC}

\begin{figure}[b]
    \centering
    \includegraphics[width=0.9\linewidth]{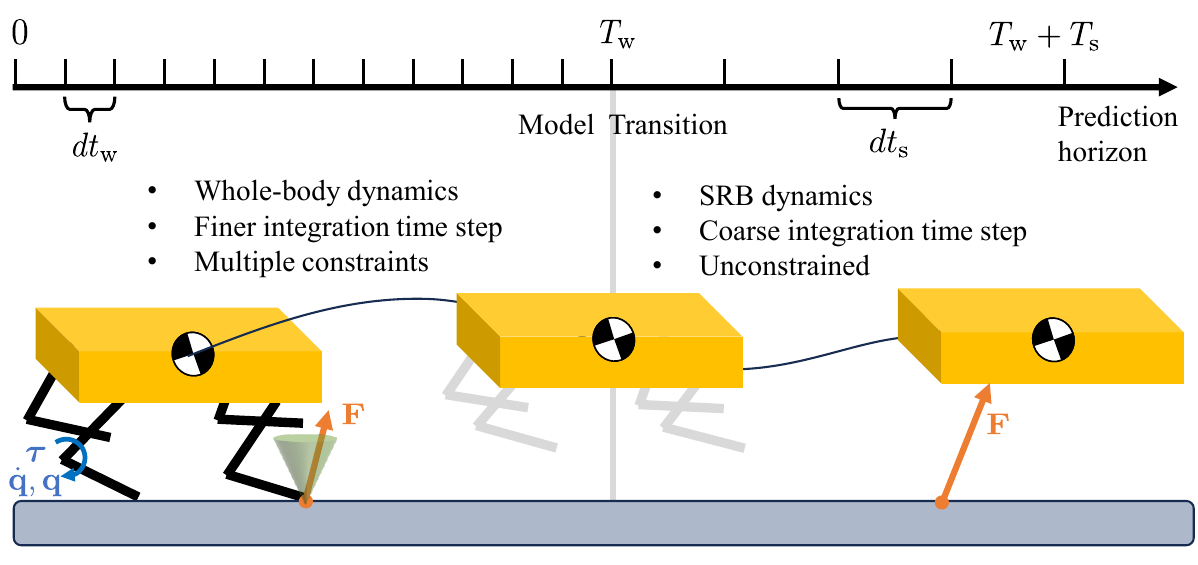}
    \caption{Illustration of the sequentially cascaded-fidelity plans along the prediction horizon.}
    \label{fig:CAFE-MPC_illustration}
\end{figure}

This section describes in detail the {\sc Cafe-Mpc} formulation. Our previous work \cite{li2021model} proposed an MHPC formulation that schedules a sequence of models with descending fidelity along the prediction horizon in each planning problem. The leading part of the plan (Fig.~\ref{fig:CAFE-MPC_illustration}) reasons about higher-fidelity models that are more consistent with the physics of the real robot. The tailing part of the plan reasons about less-expressive models to gain computational efficiency. The tailing part provides an approximation of the long-term cost to guide the leading part of the plan. With the same motivation, {\sc Cafe-Mpc} generalizes this idea of cascaded fidelity beyond dynamics modeling. Specifically, it employs a finer integration time step in the near term and a coarse time step in the long run. Further, it considers the full set of constraints in the leading plan and removes certain constraints in the tailing plan.

An overview of the {\sc Cafe-Mpc} formulation used in this work is shown in Fig.~\ref{fig:CAFE-MPC_illustration}, illustrated over a quadruped robot. We use the whole-body dynamics for the leading plan, and the single-rigid-body (SRB) dynamics for the tailing plan. The durations of the two plans are denoted by $T_w$ and $T_s$, respectively. As opposed to planar models in previous work \cite{li2021model}, these 3D representations enable more behaviors in the 3D space, for instance, a barrel roll. We use multi-resolution integration time steps, $\dt_{\text{w}} =$ 10 ms for whole-body dynamics, and $\dt_{\text{s}} =$ 50 ms for the SRB dynamics. In Section~\ref{sec:result_mhpc}, we study the effect of different $\dt_{\text{s}}$ on tracking performance and solve time. We consider a rich set of constraints in the leading whole-body plan, such as torque limits, friction constraints, etc, and leave the the tailing SRB plan unconstrained. The rest of this section describes how each planning problem is constructed, and how they are connected and are cast into a multi-phase TO problem.

\subsection{Whole-Body Plan Formulation}
\label{sec:wb_plan}
The {\sc Cafe-Mpc} formulation in this work employs whole-body dynamics with hard contacts for the higher-fidelity leading plan. We use Euler angles $\btheta \in \set{R}^3$ to represent the orientation of the floating base. Let $\c \in \set{R}^3$ be the $xyz$ position of the floating base, $\q_J\in\set{R}^{12}$ be the joint angles, $\taub\in\set{R}^{12}$ be the actuation torques, and $\q = \trans{[\trans{\btheta}, \trans{\c}, \trans{\q_J}]} \in \set{R}^{18}$ be the generalized coordinates, and $\p\in\set{R}^{12}$ be foot locations. Let $\x = [\trans{\q}, \trans{\qd}]^{\top}$ be the state variable, and $\u=\taub$ be the control variable.

\subsubsection{Cost Function}
The running cost in this work consists of three terms (1) tracking of a state reference (2) tracking of swing foot positions and velocities and (3) torque minimization, i.e.,
\begin{multline}\label{eq:rcost_wb}
    l_k =  
    \snorm{\delta\q}_{\W_q} + \snorm{\delta\qd}_{\W_{\Dot{q}}} + \\ \snorm{\S\delta\p}_{\W_{p}} + \snorm{\S\delta\Dot{\p}}_{\W_{\Dot{p}}} + \snorm{\u}_{\R_{\tau}} 
\end{multline}
where the operation $\delta \cdot$ denotes the difference between a variable and its reference. Let $s_j$ denote the swing status of the $j^{\text{th}}$ leg, then $\S\in\set{R}^{12\times12}$ is a diagonal matrix whose diagonal concatenates vectors $[s_j,s_j,s_j]$ of all four legs. The terminal cost is similarly defined but without the last term.
\subsubsection{Dynamics}
The dynamics of a quadruped robot in contact are given by
\begin{equation}\label{eq:Lagrange_dyn}
    \M(\q)\qdd + \n(\q,\qd) = \trans{\B}_{\tau}\taub + \trans{\J}_c\F_c
\end{equation}
where $\M$ is the generalized mass matrix, $\n$ collects the Coriolis, centripetal, and gravity forces, $\B_{\tau} \in \set{R}^{12\times18}$ is a constant selection matrix due to un-actuated floating base, $\F_c \in \set{R}^{3n_c}$ concatenates the ground reaction force (GRF) $\f_{EE_j} \in \set{R}^3$ of each contact foot, with $n_c$ the number of active contacts, and $\J_c \in\set{R}^{3n_c\times18}$ is the contact Jacobian. In addition to the dynamics~\eqref{eq:Lagrange_dyn}, the contact foot is often assumed to be static with the acceleration-level non-slip constraint
\begin{equation}\label{eq:nonslip_bm}
    \J_c\qdd + \Dot{\J}_c\qd = - \alpha\J_c\qd
\end{equation}
where we use the first-order Baumgarte stabilization \cite{baumgarte1972stabilization, kuindersma2016optimization} to mitigate the violation of non-slip constraint at the velocity level due to numerical integration, and $\alpha > 0$ represents the first-order Baumgarte stabilization parameter. We use $\alpha = 10$ in this work for a stabilization time constant of $1/10$ second. Further, we address the constraint~\eqref{eq:nonslip_bm} at the dynamics level, resulting in the well-known KKT contact dynamics\cite{budhiraja2018differential} 
\begin{equation}\label{eq:KKT_dyn}
    \mat{\M & \J_c^T \\ \J_c & \mathbf{0}}
    \mat{\qdd \\ -\F_{c}}
=   \mat{\B_{\tau}^T\taub - \n(\q, \qd)\\
        -\Dot{\J}_c\qd+\alpha\J_c\qd
}.
\end{equation}
To model a change in the contact mode, when a new foot touches down, the impact dynamics are similarly defined as
\begin{equation}\label{eq:KKT_impact}
    \mat{\M & \J_c^T \\ \J_c & \mathbf{0}}
    \mat{\qd^{+} \\ -\Lambda_{c}}
=   \mat{\M \qd^-\\
        \mathbf{0}
}.
\end{equation}
where the superscripts $^{+}$ and $^-$ indicating post and pre-impact event, and $\Lambda_c$ represents the impulse.
The linear systems~\eqref{eq:KKT_dyn} and \eqref{eq:KKT_impact} can be efficiently solved with Cholesky decomposition when $\J_c$ has full rank, which is performed using Pinocchio\cite{carpentier2019pinocchio} in this work. The contact dynamics~\eqref{eq:KKT_dyn} and the impact dynamics~\eqref{eq:KKT_impact} vary based on the foot contact status, resulting in a hybrid system. Assuming fixed contact sequence and timing throughout this work, we obtain the discrete-time state-space equation via explicit Euler integration 
\begin{equation}
    \xi_{k+1} = \xi_k + \mat{ \qd^{[i]}_k  \\[.5ex] \qdd^{[i]}_k}\dt_{\text{w}},
\end{equation}
where the superscript $ ^{[i]}$ follows the convention of Eq.~\eqref{eq:switchDyn} to indicate phase-dependent dynamics. It is emphasized here that $\dt_{\text{w}}$ is the integration time step for the whole-body plan. Similarly, the state-space reset map $\P_{\text{w}}$ is 
\begin{equation}\label{eq:wb_reset}
    \x^{[i+1]}_0 = \mat{\trans{(\q^{[i]^{-}})}, \trans{(\qd^{[i]^+})}}^{\top} := \P_{\text{w}}^{[i]}(\xi_N).
\end{equation}
Care must be taken here that the reset map~\eqref{eq:wb_reset} should also consider the case of taking off, where the mode-transition dynamics are trivial, with $\qd^+ = \qd^-$.
\subsubsection{Constraints}
Multiple constraints are considered in the whole-body plan. At the moment of touchdown, we require that the height of the touchdown foot be on the contact surface. Let $\p_{EE_j}$ be the foot position of the $j^{\text{th}}$ foot in world frame. Then the touchdown constraint is
\begin{equation}
    \mat{0, 0, 1} \p_{EE_j} - h_c = 0 \quad \forall j \ \text{touch down}
\end{equation}
where $h_c$ is the height of the contact surface in the world frame, $\p_{EE_j}$ is calculated via the forward kinematics. For each contact foot, the contact force needs to satisfy the friction cone constraint. In this work, we use inner pyramid approximation
\begin{equation}\label{eq:friction_cone}
     |f_{EE_j}^{x,y}| \leq \mu f_{EE_j}^z
\end{equation}
where $\mu > 0$ is the friction coefficient. Note that the Eq.~\eqref{eq:friction_cone} implies $f_{EE_j}^z \geq 0$. In addition, the torque limit, joint limit, and joint speed limit are always enforced throughout the whole-body plan
\begin{subequations}
    \begin{align}
        \taub_L &\preceq \taub \preceq \taub_U, \\
        \q_{J_L} &\preceq \q_J \preceq \q_{J_U}, \\
        \Dot{\q}_{J_L} &\preceq \Dot{\q}_J \preceq \Dot{\q}_{J_U}, 
    \end{align}
\end{subequations}
where the subscripts $L$ and $U$ indicate lower bound and upper bound, and $\preceq$ denotes element-wise inequality.

\subsection{SRB Plan Formulation}
The {\sc Cafe-Mpc} formulation in this work employs the SRB model for the lower-fidelity tailing plan. Further, the tailing plan uses coarse integration time step $\dt_{\text{s}}$ = 50 ms, and removes all constraints. For the SRB model, let $\x = [\trans{\btheta}, \trans{\c}, \trans{\Dot{\bomega}}, \trans{\Dot{\c}}]^{\top}$ be the state variable, $\u = [\trans{\f}_{EE_1},\trans{\f}_{EE_2},\trans{\f}_{EE_3},\trans{\f}_{EE_4}]^{\top}$ be the control variable (for the ground forces at the feet). Additionally, let $\bomega \in \set{R}^3$ be the angular velocity of the floating base in the body frame. 

\subsubsection{Cost Function}
The running cost consists of a tracking cost and a GRF regularization term
\begin{equation}\label{eq:rcost_SRB}
    l_k =  
    \snorm{[\delta\btheta, \delta\c]}_{\W_{\theta,c}} +  \snorm{[\delta\Dot{\btheta}, \delta\Dot{\c}]}_{\W_{\Dot{\theta},\Dot{c}}} + \snorm{\Tilde{\S}\delta\F_{EE}}_{\R_{u}} 
\end{equation}
where $\W_{\theta,c}$ are the components in $\W_q$ corresponding to $\theta$ and $\c$, and $\W_{\Dot{\theta},\Dot{c}}$ are the components in $\W_{\Dot{q}}$ corresponding to $\Dot{\btheta}$ and $\Dot{\c}$, and $\Tilde{\S} = \eye_{12} - \S$ is the contact status matrix where $\eye_{12} \in \set{R}^{12\times12}$ is an identity matrix and $\S$ denotes swing status as defined in Eq.~\eqref{eq:rcost_wb}. The terminal cost is similarly defined but excludes the regularization term.
\subsubsection{Dynamics}
The SRB dynamics is 
\begin{equation}\label{eq:SRB}
    \begin{aligned}
    \Ddot{\c} &= \sum_j^{4} \Tilde{s}_j\frac{\f_{EE_j}}{m} -\g \\
    \Dot{\bomega} &= \I^{-1}(-\bomega\times\I\bomega + \R^{\top}\sum_{j=1}^{4} \Tilde{s}_j(\p_{EE_j} - \c) \times\f_{EE_j}),
    \end{aligned}
\end{equation}
where $\I$ is the rotational inertia of the body, $\g$ is the earth gravity, $\R$ denotes the body orientation w.r.t. the world frame, $\Tilde{s}_j$ is a diagonal component of $\Tilde{\S}$ indicating the contact status of the $j^{\text{th}}$ foot. All quantities except for $\bomega$ are expressed in the world frame. Let $\T(\cdot)$ be the transformation matrix that converts angular velocity to the rate of change of Euler angles \cite{waldron2016kinematics}. With $\Dot{\btheta} = \T(\btheta)\bomega$, its time derivative, and the second equation of~\eqref{eq:SRB}, we can obtain $\,\Ddot{\!\btheta}$ as a function as the SRB state. The discrete-time state-space equation of SRB dynamics is then
\begin{equation}\label{eq:srb_ss}
    \x_{k+1} = \x_{k} + 
    \mat{
    \Dot{\btheta}^{\top} &
    \Dot{\c}^{\top} &
    \, \Ddot{\!\btheta} ^{\top} &
    \Ddot{\c}^{\top}
    }^{\top}
    \dt_{\text{s}},
\end{equation}
where it is emphasized here that $\dt_{\text{s}}$ is the integration time step for the simplified model. Note that the foot location $\p_{EE_j}$ is assumed to be known from a reference trajectory according to Raibert heuristics \cite{di2018dynamic}, it is neither part of the state nor the control.

\subsubsection{Constraints} 
The SRB planning is formulated as an unconstrained optimization problem. The idea of {\sc Cafe-Mpc} is to relax the problem constraints/cost/dynamics later in the horizon for computational efficiency over accuracy. In our case, we empirically find that removing the constraints on the tail is adequate for the highly dynamic motions we aim to produce. 

\subsection{Connecting WB Plan $\&$ SRB Plan}
The WB plan and the SRB plan are not decoupled in \cmpc, but rather are connected via a transition constraint. To differentiate their dimensionality, we use $\x_{\text{w}}$ to denote the whole-body state and $\x_{\text{s}}$ to denote the SRB state in this section. At the instance of model transition (Fig.~\ref{fig:CAFE-MPC_illustration}), the whole-body plan and SRB plan are connected via
\begin{equation}
    \x^{+}_{\text{s}} = \T_{w\rightarrow s}\P_R({\x^{-}_{\text{w}}})
\end{equation}
where the superscripts $^+$ and $^-$ denote the moment immediately before and after the model transition, $\P_R(\cdot)$ is given by Eq.~\eqref{eq:wb_reset}, and
\begin{equation}
    \T_{w\rightarrow s} = 
    \begin{bmatrix} 
    \eye_{6} & \mathbf{0}^{6\times 12} & \mathbf{0}^{6\times6} & \mathbf{0}^{6\times 12} \\
    \mathbf{0}^{6\times 6} & \mathbf{0}^{6\times 12} & \eye_{6} & \mathbf{0}^{6\times 12}
    \end{bmatrix}_{12\times36}
\end{equation}
is the state projection matrix. 
\subsection{Cast To Multi-Phase TO}
As discussed in Section~\ref{sec:MSDDP}, a new phase is determined when there is a change in one of the following (1) system dynamics (2) state or control (3) cost function (4) constraint. A rough choice of phases would be to consider the full whole-body plan as one phase and the SRB plan as another. This choice of phases, however, is not sufficiently accurate, as the contact status of each foot can likely change along the whole-body plan. The dimension of the KKT matrix of the contact dynamics depends on the number of active contacts. Therefore, the whole-body plan is further divided into multiple phases depending on foot contact status. A whole-body phase is determined when there are any foot contact changes. Since the contact schedule is known in priori, the number of whole-body phases $n_{p_w}$ as well as the start and end time of each phase can be induced given the prediction horizon $T_{\text{w}}$ of the whole-body plan. The SRB plan is considered as one single phase, since with the predetermined contact schedule, the variable $\Tilde{s}$ is simply a time-varying parameter, and the foothold locations $\p_{EE_j}$ are specified from a reference trajectory \cite{di2018dynamic}.


\begin{remark}
    We emphasize two important features of the {\sc Cafe-Mpc}. Firstly, the tailing SRB plan provides a low-rank Hessian approximation for the long-term cost to go of the whole-body model. Adding future costs to an optimal control problem empirically helps make the state more viable \cite{wieber2008viability}, provided the cost function is designed properly. An ideal case is perhaps to extend the whole-body plan with the same problem structure. This strategy, however, is computationally expensive. Relaxing the dynamics/cost/constraints later in the prediction saves computational effort, while still capturing salient features of the future plan. Secondly, the choice of the whole-body model and SRB model can be generalized to other sequences of models \cite{li2021model}. For instance, \cite{wang2021multi} employs a centroidal followed by a choice of convex model (such as linear inverted pendulum (LIP)) for humanoid walking, and \cite{khazoom2022humanoid} employs a centroidal model followed by an SRB model for humanoid balancing.
\end{remark}

\begin{remark}
    We discuss the major difference between {\sc Cafe-Mpc} and the adaptive-complexity MPC \cite{norby2022adaptive}. {\sc Cafe-Mpc} aims to robustify the whole-body MPC by adding a low-rank future cost without significantly increasing computational efforts. 
    The adaptive-complexity MPC, by contrast, focuses on adjusting the model expressiveness based on the need for the motion complexity, and, in the worst case, comes at the same computational cost as whole-body MPC. Naturally, these two ideas could be combined, for example, changing the prediction model in the tail as needed to further improve the long-term cost predictions in {\sc Cafe-Mpc}.
\end{remark}

%% file: MS_01/05_VWBC.tex
\section{Value-Based Whole-Body Controller}
\label{sec:VWBC}
The whole-body plan of {\sc Cafe-Mpc} incorporates the actuation torque ($\taub$) and the whole-body states ($\q$, $\qd$), which are executable on the robot. As a result, one strategy for WBC is to directly apply the feed-forward torque with a PD controller that regulates toward ($\q$, $\qd$). This approach essentially results in an open-loop MPC, and requires higher update frequency to account for model uncertainties. The whole-body plan, however, includes a value function approximation and a local feedback policy, which are available for free as intermediate results of MS-iLQR. An alternative approach for WBC is thus to apply the closed-loop control
\begin{equation}\label{eq:Riccati_policy}
    \pib_k(\x_k) = \u_k^* + \K_k(\x - \x^*_k)
\end{equation}
where $\K = -\inv{\Q}_{\u\u}\Q_{\u\x}$ as in Eq.~\eqref{eq:control_policy}. This approach is known as the Riccati feedback control. It smooths the actuation torque in between MPC time steps, enables improved stability, and allows for slower MPC update \cite{grandia2019feedback, mastalli2023inverse}. Despite these benefits, a well-known problem with \eqref{eq:Riccati_policy} is that the closed-loop trajectory does not necessarily satisfy certain physical constraints. This problem can be critical to the robot stability, for instance, the robot can fall when friction cone constraints are violated.

\subsection{Value-Function Based WBC}
The value-function-based WBC (VWBC) proposed in this work embeds the Riccati feedback controller~\eqref{eq:Riccati_policy} within a conventional WBC-QP formulation, thereby unifying the WBC and the Riccati feedback control. The VWBC produces a whole-body control command that is close to the Riccati controller, while at the same time satisfying all necessary physical constraints. The VWBC is formulated as a QP
\begin{subequations}\label{eq:VWBC}
\begin{IEEEeqnarray}{rl}
\min_{\taub,\qdd,\blambda} \ & Q_k(\x-\x^*_k, \taub-\taub^*_k) \ \label{eq:vwbc_cost}\\
\text{s.t.} \ &  \M(\q)\qdd + \n(\q,\qd) = \trans{\B}_\tau \taub + \trans{\J}_c\F_{c} \label{eq:eom} \\
 & \J_j\qdd + \Dot{\J}_j\qd = - \alpha\J_j\qd, \quad \forall j \in \mathcal{C} \label{eq:nonslip}\\
 & \taub_L \preceq \taub \preceq \taub_U, \label{eq:torque_limit} \\
 & f_{EE_j}^z \geq 0, \quad \forall j \in \mathcal{C} \label{eq:unilateral}\\
 & |f_{EE_j}^{x,y}| \leq \mu f_{EE_j}^z, \quad \forall j \in \mathcal{C} \label{eq:friction}
\end{IEEEeqnarray}    
\end{subequations}
where $Q_k(\cdot,\cdot)$ is the action-value function defined in Eq.~\eqref{eq:Q_approx}, $k$ is the time index in the whole-body plan of {\sc Cafe-Mpc}, the Eq.~\eqref{eq:eom} is the whole-body contact dynamics as in Eq.~\eqref{eq:Lagrange_dyn}, $\mathcal{C}$ denotes the set of active contacts. Eq.~\eqref{eq:nonslip} represents the acceleration-level non-slipping constraint with first-order Baumgarte stabilization. Eqs.~\eqref{eq:torque_limit} - \eqref{eq:friction}, respectively, represent the torque limits, the unilateral constraint, and the linearized friction-cone constraint with an inner pyramid approximation. 

The partial derivatives of $Q_k(\cdot, \cdot)$, the (sub)optimal state-control pair $(\x_k^*,\u_k^*)$, and the contact status $\mathcal{C}$ are obtained from the whole-body plan of {\sc Cafe-Mpc}, and are queried at a time that is the closest to the current low-level control tick. Minimizing $Q_k$ will encourage the resulting solution to stay close to the optimal control policy~\eqref{eq:control_policy}. Since $\Quu$ in Eq.~\eqref{eq:Q_approx} is guaranteed (via regularization) to be positive definite, the resulting QP~\eqref{eq:VWBC} is strictly convex.
    
The VWBC~\eqref{eq:VWBC} is a generalization of the Riccati controller~\eqref{eq:control_policy}. This can be seen by removing all the constraints in~\eqref{eq:VWBC}. The problem~\eqref{eq:VWBC} then becomes an unconstrained optimization problem, the solution of which is Eq.~\eqref{eq:control_policy}. Thus, the VWBC seeks a control signal that is the closest to the Riccati controller~\eqref{eq:control_policy} but with all necessary constraints satisfied. In other words, the VWBC can be considered as the Riccati controller disguised in a QP, thus enjoying the benefit of feedback stabilization, while preventing the resulting solution from being too aggressive to violate the constraints. The VWBC is similar in spirit as~\cite{kuindersma2014efficiently, chignoli2020variational}, except that the value functions in~\cite{kuindersma2014efficiently, chignoli2020variational} are obtained via LQR designed for simplified models.

The VWBC avoids additional cost tuning beyond the {\sc Cafe-Mpc}. The action-value function $Q_k(\cdot,\cdot)$ marries the VWBC to the {\sc Cafe-Mpc}. Thus, one can focus on the cost design for the {\sc Cafe-Mpc}, and leave {\sc Cafe-Mpc} to regulate the VWBC. This is an advantage over the conventional OSC-QP, since conventional OSC-QP concatenates multiple tasks specified in the operational space, such as center-of-mass (CoM) tracking, swing feet tracking, torque regularization, etc. As a result, nontrivial tuning is often unavoidable to balance the relative importance of each task. Further, these tasks are often generated independently using separate planners that are based on simplified models. For instance, CoM trajectories can be generated using LIP model \cite{wieber2006trajectory, kuindersma2014efficiently, apgar2018fast, xiong20223}, SRB model \cite{di2018dynamic, bledt2018cheetah, kim2019highly}, and Centroidal dynamics \cite{kuindersma2016optimization, grandia2023perceptive}. Swing foot trajectories are often generated by interpolating predicted foot placements using Bezier polynomials \cite{di2018dynamic, bledt2018cheetah, grandia2023perceptive}, where the foot placements are obtained with Raibert heuristics. The loss of whole-body information can potentially produce a plan that is not trackable by the low-level controller \cite{focchi2016robot}. The VWBC, by contrast, overcomes the above issues, and avoids any additional meticulous tuning beyond the MPC.


The QP~\eqref{eq:VWBC} can further be warm started using solutions of {\sc Cafe-Mpc}, and thus requiring fewer solver iterations. The whole-body plan of {\sc Cafe-Mpc} incorporates $\taub_k^*, \q_k^*, \qd_k^*, \F_{c,k}^*$. We obtain $\qdd_k^* = (\qd_{k+1}^* - \qd_k^*)/\dt_{\text{w}}$, and use $\taub_k^*, \qdd_k^*, \F_{c,k}^*$ as an initial guess for the QP~\eqref{eq:VWBC}. As will be shown in Section~\ref{sec:result_vwbc}, only one iteration is needed most of the time for a bounding gait with an active set method.

%% file: MS_01/06_Implementation.tex
\section{Implementation Details}
\label{sec:implementation}
In this section, we discuss a few implementation details, including reference generation, offline TO design for the barrel roll, motion composition of running barrel roll, and other engineering details that are important to maintain fast computation and robust hardware execution. For regular locomotion skills such as bounding, pacing, etc, we use sparse kinematic references that are based on integrating the twist and Raibert heuristics. For the barrel roll and any composed motions that incorporate the barrel roll, the reference trajectories are obtained via offline TO that covers all degrees of freedom.


The contact patterns and timings are heuristically determined by observing the gaits of similar-size quadruped animals. The contact patterns are represented by a switched system formulation, i.e., an ordered sequence of phases. Each phase is associated with a start time and an end time. The change of phase is determined by a take-off event and/or a touch-down event of any legs. Let FR, FL, HR and HL represent the stance status of the front right, front left, hind right, and hind left legs, respectively. Let FT represent a flight phase, and FS represent a full-stance phase. Then the contact sequence of one gait cycle of a bounding gait is represented by $\{$HL-HR, FT, FL-FR, FT$\}$, with the associated timings $\{[t^{[1]}_s, t^{[1]}_e], [t^{[1]}_e, t^{[2]}_e], [t^{[2]}_e, t^{[3]}_e], [t^{[3]}_e, t^{[4]}_e]\}$. Such switched-system representation makes formulating the multi-phase TO problem~\eqref{eq:c-multi-phase} straightforward. While heuristics is used here to determine the contact patterns, an alternative approach is to employ a contact planner, either via offline design \cite{posa2014direct, winkler2018gait} or via online re-planning \cite{kuindersma2016optimization, tassa2012synthesis}.

\subsection{Heuristic Locomotion Reference}
\label{sec:loco_reference}

The kinematic reference for regular locomotion skills is sparse and is designed via simple heuristics, consisting of the CoM trajectory, the foot placements, and the swing foot trajectory. The users provide commands that include horizontal velocities $v_x$ and $v_y$, height $z$, and yaw rate $\Dot{\theta}_z$. The horizontal positions $x,y$, and the yaw angle $\theta_z$ are obtained via integrating the corresponding velocity components. The rest of CoM states are set to zero. The reference foot placements are obtained via Raibert heuristics \cite{di2018dynamic, bledt2018cheetah}, but are clamped to be within a bounded box about the corresponding hip so that they are kinematically reachable. The foot placements are interpolated with cubic Bezier polynomials \cite{di2018dynamic, bledt2018cheetah}, from which the swing foot positions and velocities are obtained. All components of the kinematic reference are used in Eq.~\eqref{eq:rcost_wb} to design the cost of the whole-body plan. The CoM reference is used in Eq.~\eqref{eq:rcost_SRB} to design the cost of SRB plan, as well as normal GRF references that are obtained by averaging the total weight of the robot over the number of active contacts.

\subsection{Offline Barrel Roll TO}
While a sparse kinematic reference is sufficient for regular locomotion skills, we empirically find that richer information about leg coordination is helpful to quickly synthesize complex motions online such as barrel roll given the real-time constraints. As such, we solve a TO problem offline to obtain a whole-body reference trajectory for an in-place barrel roll starting on all fours. We then let the {\sc Cafe-Mpc} perform the fine-tuning online to account for the model mismatch or mismatch in initial motion or contact configuration. Details of this offline design process can be found in the Appendix.

\subsection{Engineering Details}
Engineering implementations are important for successful and robust executions on robot hardware. In this section, we review some of these implementation details in terms of {\sc Cafe-Mpc} configuration, warm-start strategy, policy-lag compensation, evaluation of dynamics and its partials, etc.

\begin{figure}
    \centering
    \includegraphics[width=0.8\linewidth]{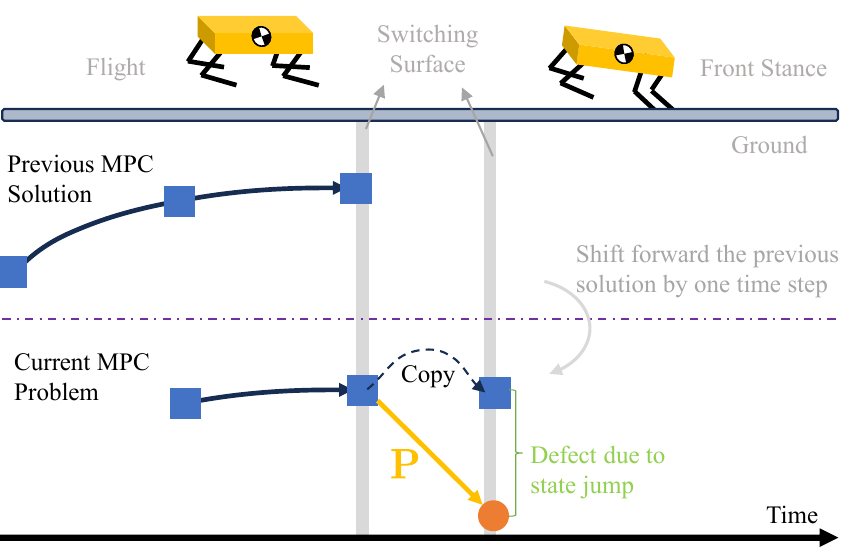}
    \caption{Illustration of large defect when a new phase involving state jumps is added to the current MPC problem. In warm-starting the current MPC problem, the previous MPC is shifted forward by one step. The last state of the previous MPC solution is duplicated to initialize the newly created state, which produces a large defect due to the state jump. The illustration example shows when the MPC moves out of a flight phase, and adds in a new stance phase.}
    \label{fig:cafempc-warmstart}
\end{figure}

The {\sc Cafe-Mpc} is configured to use fixed prediction horizons (i.e., whole-body plan horizon $T_w$, and SRB plan horizon $T_s$ in Fig.~\ref{fig:CAFE-MPC_illustration}). As discussed in Section~\ref{sec:MHPC}, the SRB plan is designed to have a single phase while the whole-body plan is constructed to span multiple phases. As the {\sc Cafe-Mpc} shifts forward, the phases involved in the whole-body plan may vary as the most recent contact status moves out of the horizon, and the upcoming contact status moves in (as illustrated in Fig.~\ref{fig:cafempc-warmstart}). In other words, the {\sc Cafe-Mpc} may entail dynamic removal of old phases and addition of new phases.

It is well accepted that for fixed-horizon MPC, the optimization problems between two control ticks are similar. Thus, a common approach to save computational effort is to warm start the current MPC problem using the solution from the previous MPC problem. In constructing the current MPC problem, the previous MPC solution is shifted forward by a certain amount of time steps, and new decision variables of proper sizes are appended to the shifted trajectory so that the prediction horizon remains invariant. With a multiple-shooting-based solver, the new state variables are initialized with the last state in the previous MPC solution. Care must be taken, however, on this warm-start strategy for multi-phase problems~\eqref{eq:c-multi-phase} subject to state jumps (e.g., impact \eqref{eq:KKT_impact}) as in {\sc Cafe-Mpc}. When a new phase is added, this scheme can produce a large defect due to the state jump (illustrated in Fig.\ref{fig:cafempc-warmstart}), potentially requiring more iterations to converge. We propose an adaptive scheme to address this issue. If no new phases are created or the new phase does not create any state jumps (e.g., stance to flight), we keep the appended states as shooting states and use the same warm-start scheme. Otherwise, the appended states are treated as roll-out states, the defect of which is identically zero. To remind the readers, a shooting state refers to an independent state that can be initialized by the user, whereas a roll-out state is a dependent variable on controls and previous states, as is detailed in Section~\ref{sec:MSDDP}. 

We solve the first {\sc Cafe-Mpc} problem to convergence. For all subsequent {\sc Cafe-Mpc} problems, the MS-iLQR solver is terminated with either maximum CPU time (18 ms) or a maximum number of iterations (4), whichever is reached first. We use Pinocchio\cite{carpentier2019pinocchio} for efficient calculation of the whole-body dynamics and its analytical derivatives, and use CasADi\cite{Andersson2019} for SRB dynamics and its derivatives. All gradient information (dynamics, costs, and constraints) along the trajectory are computed in parallel with 4 threads using OpenMP.

While the VWBC runs on the robot hardware at 500 Hz, the {\sc Cafe-Mpc} is executed on a separate computer at 33 to 50 Hz depending on the motion task. The communication between them is carried out via LCM\cite{5649358}. To account for the policy lag due to MPC solve time and communication latency, a sub-trajectory of size six in the front of the MPC solution is sent to the low-level VWBC controller. The VWBC then finds a solution that is closest to the current time. In addition, since the VWBC plays a role of Riccati feedback controller and executes at a significantly higher rate than {\sc Cafe-Mpc}, it further also minimizes the effect of policy lag.

%% file: MS_01/07_Result.tex
\section{Results}
\label{sec:results}
The performance of the proposed motion control framework is benchmarked on the MIT Mini Cheetah with a set of regular locomotion skills and highly dynamic barrel rolls. The benchmark is carried out in both a highly-fidelity dynamics simulator and on the robot hardware. The simulation and the hardware experiments share the same control structure, where the state estimation and VWBC run at 500 Hz in the main thread, and the {\sc Cafe-Mpc} runs on a separate computer (process). For regular locomotion skills, including dynamic bounding, it is sufficient to run {\sc Cafe-Mpc} at 33 Hz. For highly dynamic barrel roll, the {\sc Cafe-Mpc} runs at 50 Hz. Analysis of the {\sc Cafe-Mpc} is discussed in Section~\ref{sec:result_mhpc}. Results of VWBC are reported in Section~\ref{sec:result_vwbc}. Lastly, we compare {\sc Cafe-Mpc} to the SRB MPC controller developed in previous work \cite{li2022versatile} in terms of their capabilities to perform a barrel roll. The comparison results are discussed in Section~\ref{sec:result_br}.

\subsection{{\sc Cafe-Mpc}}
\label{sec:result_mhpc}
\begin{table}[b]
    \centering
    \caption{Summary of model schedules used for {\sc Cafe-Mpc} performance benchmark. Unit (seconds).}
    \resizebox{\columnwidth}{!}{
    \begin{tabular}{|c|c|c|c|c|c|c|c|c|c|}
        \hline
         &  $\as_0$ & $\as_1$ & $\as_2$ & $\as_3$ & $\as_4$ & $\as_5$ &$\as_{10}$ & $\as_{15}$ & $\as_{20}$\\
         \hline
        $T_{\text{w}}$ & 0.25 & 0.25 & 0.25 & 0.25 & 0.25 & 0.25 & 0.25 & 0.25 & 0.25\\
        \hline
        $T_{\text{s}}$ & 0.00 & 0.10 & 0.20 & 0.30 & 0.40 & 0.5 & 1.0 & 1.5 & 2.0\\
        \hline
    \end{tabular}}
    \label{tab:as}
\end{table}
We investigate the performance of {\sc Cafe-Mpc} in terms of its tracking capabilities and solve times under varying model schedules and different integration time steps. As mentioned in Section~\ref{sec:MHPC}, a model schedule $\as = (T_{\text{w}}, T_{\text{s}})$ specifies the whole-body prediction horizon $T_{\text{w}}$ and the SRB prediction horizon $T_{\text{s}}$. Table~\ref{tab:as} summarizes the model schedules used in the {\sc Cafe-Mpc} benchmark test, where $T_{\text{w}}$ is fixed at 0.25~s, and $T_{\text{s}}$ is varied from 0 to 2.0~s. We benchmark the {\sc Cafe-Mpc} performance on four common locomotion skills, trotting, pacing, bounding, and pronking. The VWBC is used as the whole-body control scheme consistently for all tasks and model schedules. Comparisons of different whole-body control techniques are discussed in the next section. The benchmark here advances our previous work \cite{li2021model} to 3D quadruped robots, multi-resolution integration time steps, hardware demonstration, more practical real-time MPC applications, and the more robust multiple-shooting solver.

\subsubsection{Simulation Results}
\begin{figure}[t]
    \centering
    \includegraphics[width=0.9\linewidth]{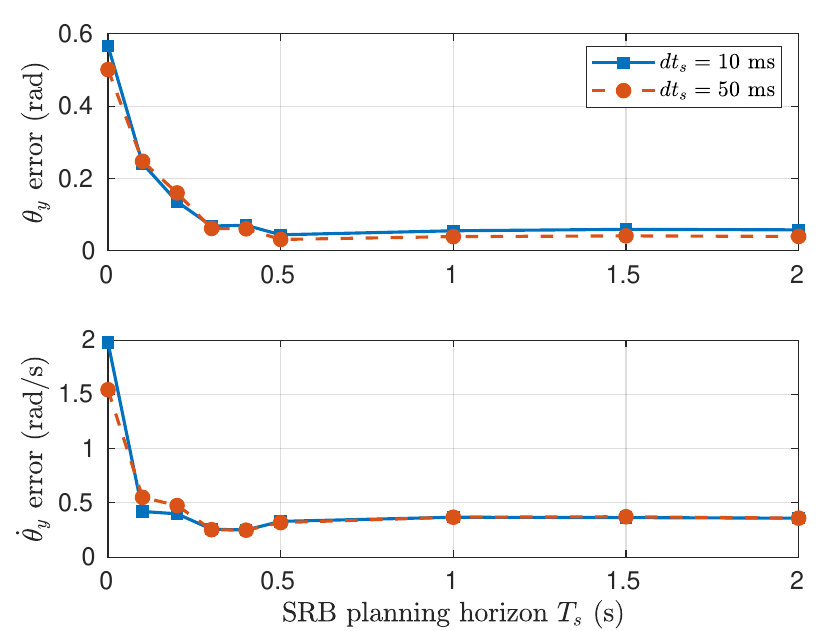}
    \caption{Tracking performance of {\sc Cafe-Mpc} under varying SRB planning horizons and different SRB integration time steps with a bounding gait on simulated Mini Cheetah. Tracking performance is measured here using RMS error on pitch $\theta_y$ (top) and pitch rate $\Dot{\theta}_y$ (bottom).}
    \label{fig:track_res_bound}
\end{figure}

Figure~\ref{fig:track_res_bound} reports the tracking performances for a bounding gait obtained in simulation. The model schedules in Table.~\ref{tab:as} are tested as well as two SRB integration timesteps, i.e., $\dt_{\text{s}}$ = 10 ms and $\dt_{\text{s}}$ = 50 ms. The tracking performances are measured using the root-mean-square (RMS) errors of the pitch angle $\theta_y$ and the pitch rate $\Dot{\theta}_y$. With $T_s = 0$~s, the \cmpc becomes a whole-body MPC, and attains the worst tacking performance. As the horizon of the tailing SRB plan grows, the tracking errors are significantly reduced ($\% $ reduction on $\theta_y$ and $\%$ reduction on $\Dot{\theta}$) until $T_s = 0.5$ s, after which the performance enhancement is minimal. In fact, with $T_s=0$ s, the large pitch variation eventually results in robot falling over. This improvement is expected. The tailing SRB plan informs the robot of a longer-term objective. Even though the long-term objective is formulated in a lower-dimensional space, it helps improve the whole-body plan in the near term, so that the robot is further-sighted and better prepared for the future goal. The lack of further performance improvement beyond $T_s = 0.5$ s is justifiable, and it is likely a result of the trailing SRB plan already serving as a reasonable indication of the long-term goal. This analysis is aligned with the results on disturbance rejection obtained in previous work \cite{li2021model}, but is benchmarked in a high-fidelity simulator herein and using more practical real-time MPC implementation. In addition, Fig.~\ref{fig:track_res_bound} shows that the tracking performance enhancement is observed for both $\dt_{\text{s}}$ = 10 ms and $\dt_{\text{s}}$ = 50 ms, and the performance differences arising from the two integration timesteps are not phenomenal. This result demonstrates that employing a fine integration timestep, later in the prediction horizon, is not critical to affecting the tracking performance. As we will show next, using a coarse integration time step is helpful to enhance the computational speed.
\begin{figure}[t]
    \centering
    \includegraphics[width=0.9\linewidth]{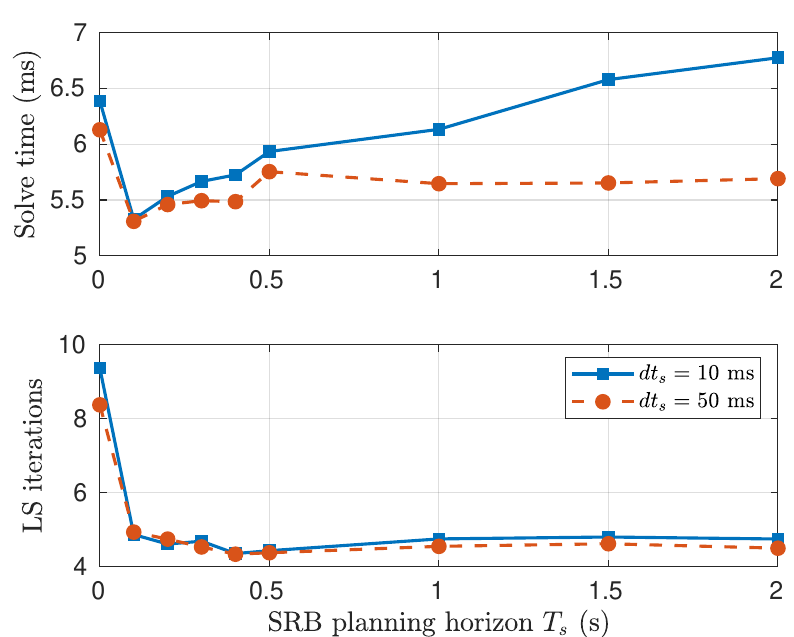}
    \caption{Solve time statistics of {\sc Cafe-Mpc} under varying SRB planning horizons and different SRB integration time steps. Top: average solve time per MS-iLQR iteration. Bottom: average number of line search (LS) iterations per MPC step.}
    \label{fig:solve_time}
\end{figure}

\begin{table}[b]
    \centering
    \caption{Average computation time (unit: ms) per MS-iLQR iteration under different testing model schedules and locomotion skills using $\dt_{\text{s}}=50$ ms}
    \begin{tabular}{|c|c|c|c|c|c|}
        \hline
         & $\as_0$ & $\as_1$ & $\as_2$ & $\as_3$ & $\as_4$\\       
        \hline
        Pace & 5.4143  &  4.8653  &  5.0176  &  4.9466  &  4.8907\\
        \hline
        FlyTrot & 5.2323  &  5.1333  &  5.0770  &  5.1805 &  5.0575 \\
        \hline
        Pronk & 5.2313  &  5.1079  &  5.0574  &  5.1948  &  5.1342\\
        \hline
    \end{tabular}    
    \label{tab:time_per_iter}
\end{table}

Figure.~\ref{fig:solve_time} depicts the solve time statistics associated with each model schedule in Table.~\ref{tab:as}, and with the two integration schemes aforementioned. The solve time per MS-iLQR iteration is measured, and is averaged over all MPC control ticks for each motion. The number of overall line search (LS) iterations at each MPC update is counted, and is averaged over all MPC control ticks. We alert the readers here that in each MPC control tick, a multi-phase TO problem~\eqref{eq:c-multi-phase} is solved. With $\dt_{\text{s}}$ = 10 ms, the solve time drops down at $T_s$ = 0.1~s, and increases afterwards. The increase in solve time is reasonable as the problem size grows with $T_s$. The decrease in solve time is accounted for by the drop in the number of LS iterations, which is likely because the enhanced horizon results in more robust feedback gains that enable more aggressive step sizes for line search. This similar behavior is also observed in the case of $\dt_{\text{s}}$ = 50 ms. Nevertheless, the solve time associated with $\dt_{\text{s}}$ = 50 ms exhibits a flat profile after $T_s$ = 0.1~s. This phenomenon is counter-intuitive at first sight. However, we note that with $\dt_{\text{s}}$=50 ms, increasing $T_s$ by 0.5 s only adds 10 more steps in the problem~\eqref{eq:c-multi-phase}. The computation cost due to the small increment of problem size might be on the same order of magnitude as the run-time variation of the computation hardware. Regardless, the solve times associated with the coarse integration timestep are in general less than those with the finer timestep, and this difference tends to increase as the planning horizon grows. Given these reasons, a coarse integration timestep is favored in the tailing SRB plan. Table.~\ref{tab:time_per_iter} shows the solve times per MS-iLQR iteration under $\as_0\sim\as_4$ for other locomotion gaits, further demonstrating that elongating the tailing SRB plan with coarse timestep does not necessarily increase solve times.

\subsubsection{Hardware Results}
\begin{figure}[t]
    \centering
    \includegraphics[width=0.9\linewidth]{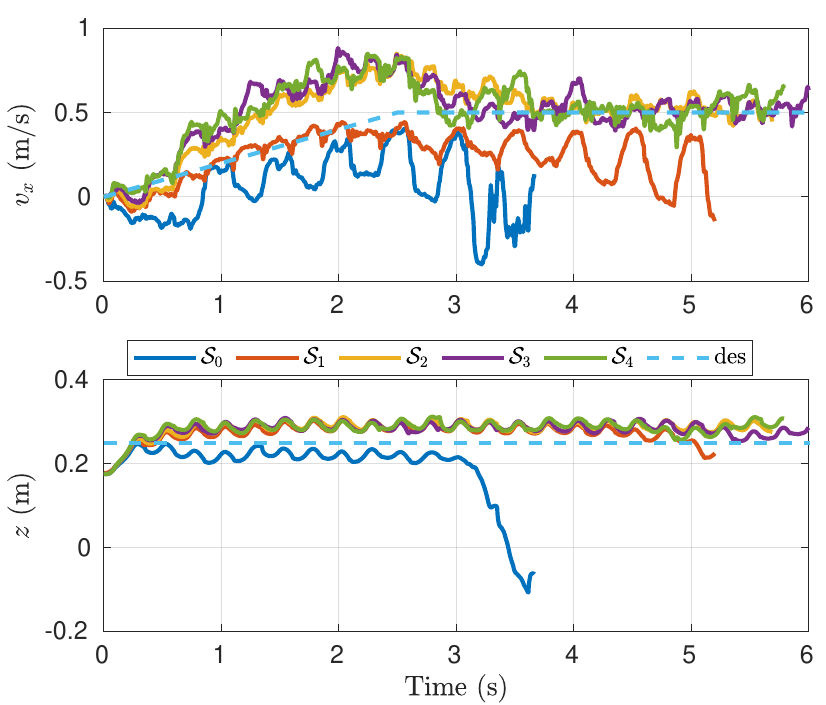}
    \caption{Tracking performance of {\sc Cafe-Mpc} under different testing model schedules evaluated with a trotting gait on Mini Cheetah hardware. } 
    \label{fig:track_res_pace}
\end{figure}

To further examine the performance of \cmpc as observed in Fig.~\ref{fig:track_res_bound}, we conduct a series of similar experiments on the Mini Cheetah hardware with a trotting gait. Rather than testing all model schedules (Table.~\ref{tab:as}) as in the previous section, we focus on $\as_0\sim\as_4$ since the most significant error reductions are observed in this range in Fig.~\ref{fig:track_res_bound}. Further, we use $\dt_{\text{s}}$ = 50 ms for the SRB plan. Figure~\ref{fig:track_res_pace} depicts the tracking results in terms of forward velocity ($v_x$) and the body CoM height ($z$). The solid lines represent the actual states of the robot under different model schedules, and the dashed line represents the desired states. With $\as_0$, the robot gradually deviates from the desired trajectory, and eventually falls down after walking for a few steps. The robot can move longer with $\as_1$ but still eventually falls down. By further elongating the SRB plan, stable trotting is achieved, and the tracking performance is improved until $\as_2$. The lack of further performance improvement is similarly observed in Fig.~\ref{fig:track_res_bound}, and is likely because the trailing SRB plan already serves as a reasonable indication of the long-term goal.

\begin{figure*}[b]
    \setcounter{figure}{9}
    \centering
    \includegraphics[width=0.9\textwidth]{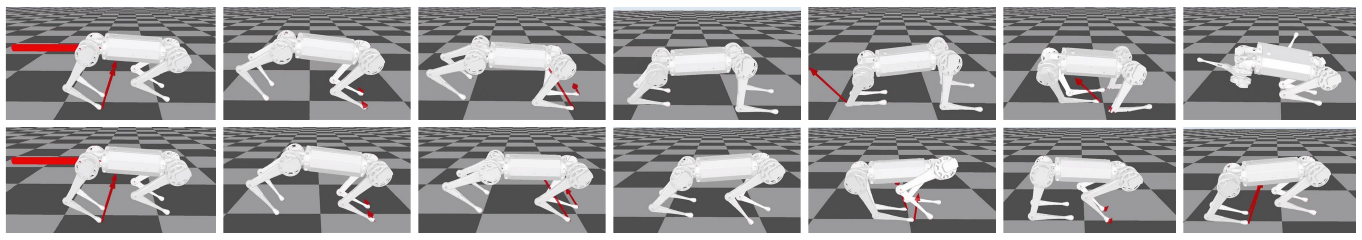}
    \caption{Time-series snapshots of Mini Cheetah bounding under forward velocity strong disturbances (1.5 m/s) in simulation using {\sc Cafe-Mpc}, and different whole-body control schemes. Top: Riccati feedback controller. Bottom: proposed VWBC.}
    \label{fig:vwbc_friction}
\end{figure*}

\subsection{VWBC}
\label{sec:res_vwbc}
\begin{figure}
    \setcounter{figure}{8}
    \centering
    \includegraphics[width=1.0\linewidth]{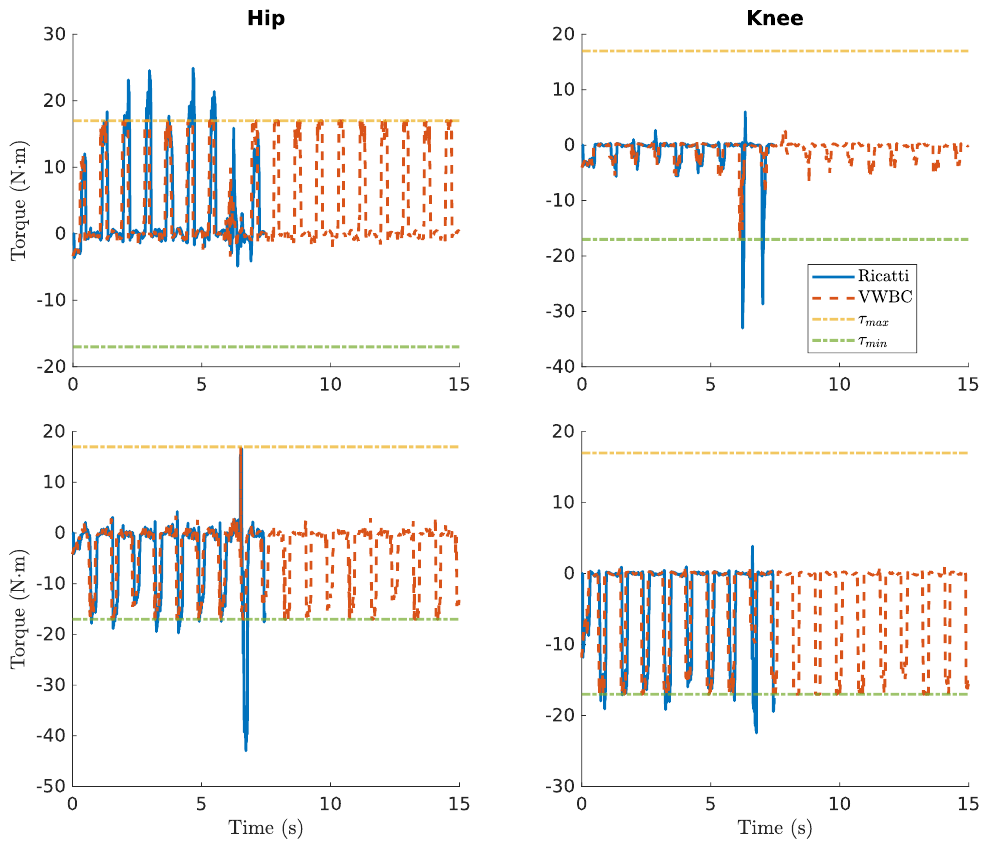}
    \caption{Hip and knee torques of the front right leg (top) and the back right leg (bottom) obtained in the disturbance rejection test. The blue lines stop at ~7.5 s due to the robot falling out of the push disturbance. $\tau_{max}$ and $\tau_{min}$ indicate the upper and lower torque limits.}
    \label{fig:vwbc_torque}
\end{figure}
\label{sec:result_vwbc}

In this section, we study the performance of the VWBC, and compare it with the performance of the Riccati controller. To remind the readers of the differences, the Riccati controller controls robots with the feedforward-feedback controller~\eqref{eq:control_policy} resulting from the MS-iLQR solver. The VWBC solves a value-based QP \eqref{eq:VWBC} that embeds~\eqref{eq:control_policy}, but with all necessary constraints satisfied. Toward that regard, we benchmark the differences with a disturbance rejection test, and are mainly concerned with constraint satisfaction and disturbance recovery. We use qpOASES \cite{Ferreau2014} as the QP solver. The whole-body plan of {\sc Cafe-Mpc} offers an initial guess for the valued-based QP without additional computational cost. We compare the QP solve times with and without using this initial guess.

\subsubsection{Disturbance Recovery}
The disturbance rejection test was conducted with the same bounding gait as in Section~\ref{sec:MHPC}. We change the body velocity by a total of 1.5 m/s over 50 ms, which corresponds to an effective external for of 270 N over the period. Figure.~\ref{fig:vwbc_torque} depicts the joint torques of the front right and back right legs as well as their upper and lower limits. It demonstrates that the Riccati controller sometimes violates the torque constraints, especially after the push disturbance, whereas the VWBC consistently satisfies the torque constraints. One may argue that a simple clamping technique can work here, and question the need for the proposed technique. This argument is true in this case, and clamping torque is sufficient to prevent control saturation. However, the VWBC provides a systematic way of balancing optimality (local Q function) and constraint satisfaction. Further, some other constraints such as friction that cannot be clamped are important to robot stability. To demonstrate, Fig.~\ref{fig:vwbc_friction} shows time-series snapshots of the robot after the push disturbance. With the Riccati controller, the robot has a non-trivial slipping that eventually disablizes the robot. With the VWBC, the slipping is slight, and the robot quickly recovers a stable contact in one gait cycle.

\subsubsection{QP Solve Time}
\begin{figure}[t]
    \setcounter{figure}{10}
    \centering
    \includegraphics[width=0.8\linewidth]{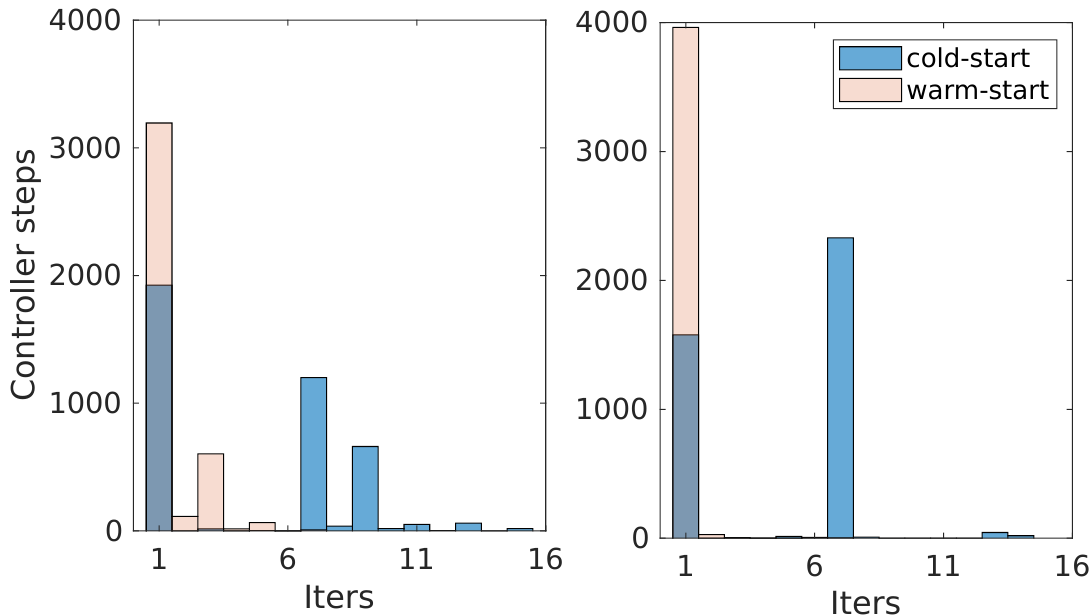}
    \caption{Histogram of the number of iterations for all controller steps with and without warm start. Left: bounding gait. Right: pacing gait.}
    \label{fig:qp_time}
\end{figure}
We compare the QP solve times (Fig.~\ref{fig:qp_time}) with and without warm starts on a bounding gait and a pacing gait without applying any disturbances, in terms of the number of solve iterations\footnote{In qpOASES, the concept of working number of recalculations (nWSR) is used. The number of iterations here is nWSR + 1.}. The number of QP iterations at every control tick (500 Hz VWBC in Fig.~\ref{fig:overview}) are collected. For bounding, the VWBC with initial guess only requires one iteration to solve the QP~\eqref{eq:VWBC} for $79.85\%$ of all times, and no more than three iterations for $97.78\%$ of all times. For pacing, $99.08\%$ of the QPs are solved with one iteration. This speedup shows a clear benefit of the proposed {\sc Cafe-Mpc} + VWBC scheme.

\subsection{Barrel Roll}
\label{sec:result_br}
\begin{figure*}[b]
    \setcounter{figure}{13}
    \centering
    \includegraphics[width=0.9\textwidth]{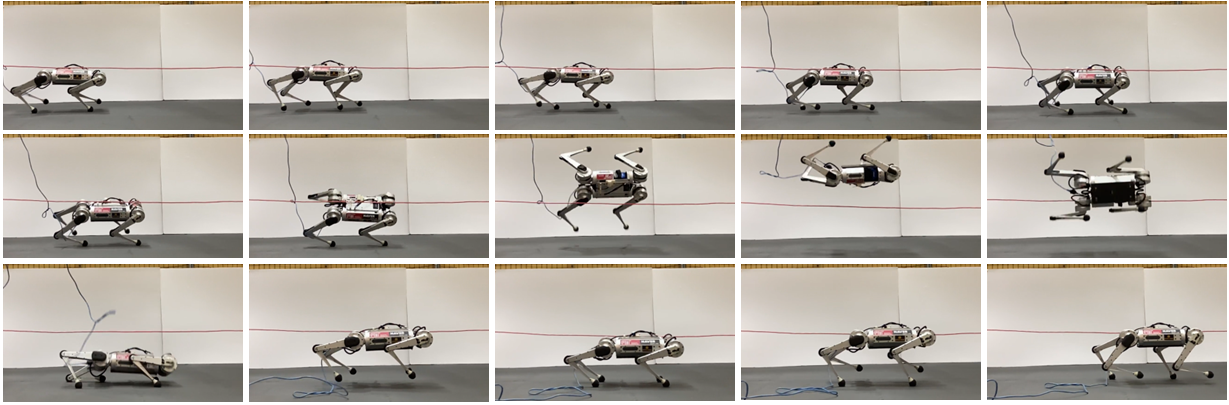}
    \caption{Time-series snapshots of the MIT Mini Cheetah performing mid-run barrel roll. The robot runs at 0.8 m/s with a pacing gait, decelerates and performs a barrel roll over a 0.4 m cable, hops for one step, and accelerates to 0.8 m/s with the pacing gait. All motions and transitions are synthesized online in real time. The proposed controller is sufficiently powerful so that no intermediate full stance is needed before the barrel roll.}
    \label{fig:flop_snapshots}
\end{figure*}
To investigate the capability of the proposed approach, we apply the {\sc Cafe-Mpc}+VWBC scheme to accomplish three tasks that each involve a highly dynamic barrel roll. (1) In the first task, the robot is commanded to execute an in-place barrel roll followed by a pacing gait. We then compare the proposed approach and a conventional MPC approach in terms of their capabilities to accomplish this task. (2) In the second task, we make the motion a bit more challenging. The robot is commanded to perform a barrel roll in the middle of a running locomotion gait. An intermediate full-stance phase is used before the barrel roll to help gain stability. Two locomotion gaits (trotting and pacing) are tested. (3) In the third task, we make the robot imitate a human athlete performing a Fosbury flop. This task is more challenging than the previous tasks as no intermediate full stance is employed before the barrel roll. We repeat this task multiple times for the reliability test. We discuss the task (1) and (3) in detail in this manuscript, and encourage the readers to check the results of task (2) in the accompanying video.

\begin{figure}[t]
    \setcounter{figure}{11}
    \centering
    \includegraphics[width=0.95\linewidth]{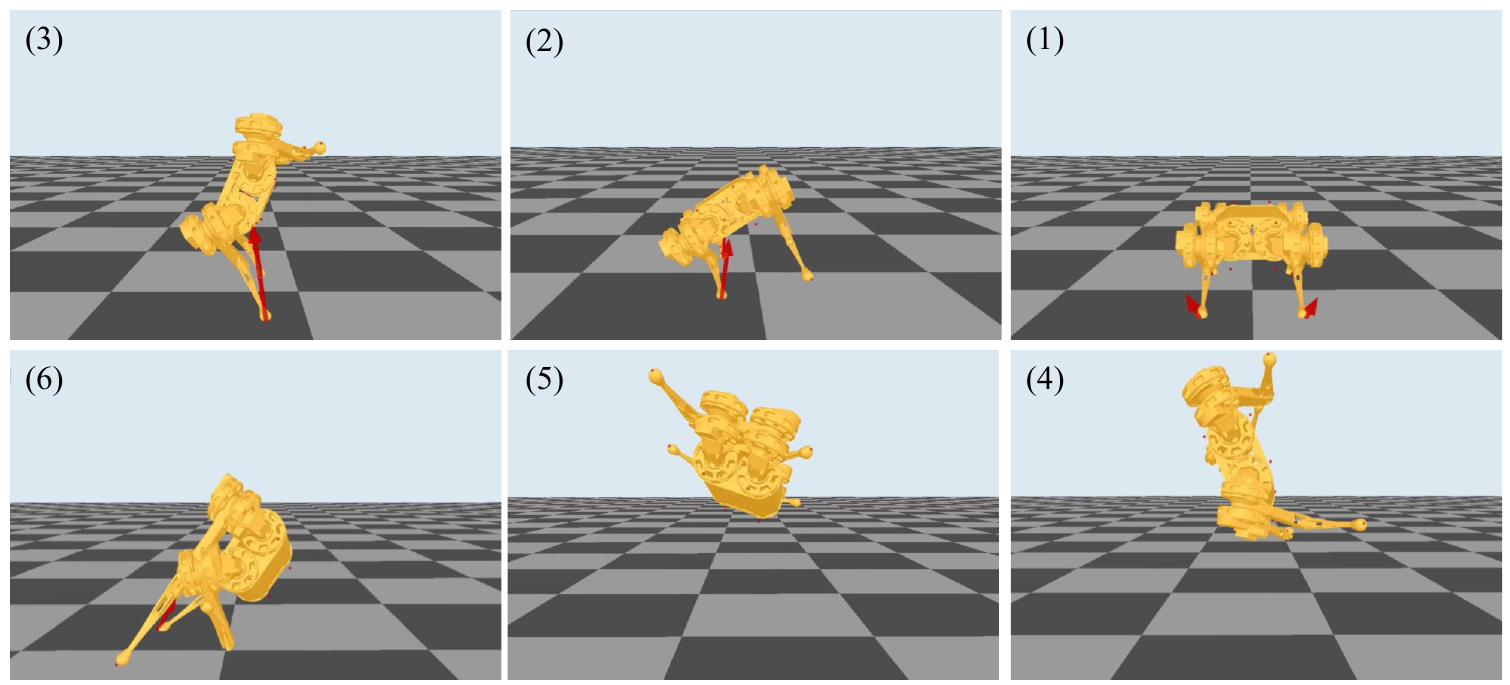}
    \caption{Time-series snapshots of in-place barrel roll using the HKD-MPC. The robot fails to accomplish the barrel-roll task because the swing-legs motions are not coordinated with the body motion.}
    \label{fig:br_hkdmpc}
\end{figure}

References for the barrel roll are generated via the offline TO as discussed in Section~\ref{sec:br_reference}. All three tasks share the same barrel-roll reference, though it is designed assuming taking off in place. References for the locomotion gaits are obtained from a similarly constructed offline TO, but the objective is to track a kinematic trajectory as introduced in Section~\ref{sec:loco_reference}\. The resulting references capture rich information in the joint space, enabling the {\sc Cafe-Mpc} to use one single cost function across all tasks. The composed reference trajectory for each task is obtained by simply connecting the two types of references in proper order without special care on motion transitions. 
\begin{figure}
    \centering
    \includegraphics[width=0.98\linewidth]{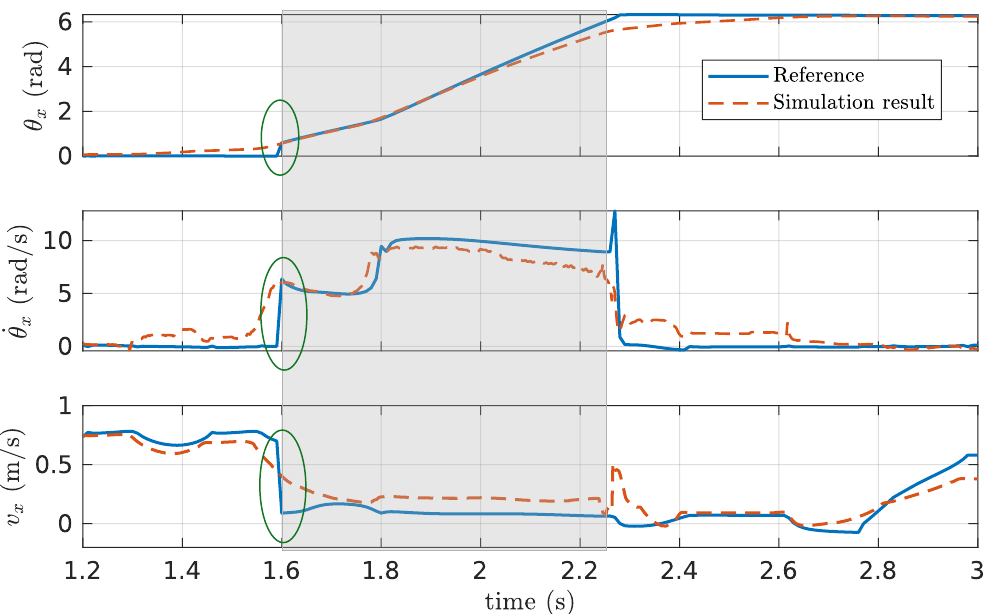}
    \caption{Offline-composed running barrel roll reference (solid blue) and the executed trajectory (dashed red) on the Mini Cheetah in simulation. The circled areas represent discontinuities when composing the barrel roll reference and the pacing reference. The grey area indicates the barrel roll.}
    \label{fig:online_smoothing}
\end{figure}

Figure.~\ref{fig:inplace-br-intro} shows time-series snapshots of executing task (1) on the Mini Cheetah hardware. The hopping step after barrel roll and before pacing is introduced by design to mimic a gymnast who often takes a step to gain balance after large aerial rotations. The successful hardware execution demonstrates the capability of the proposed framework on 
online fine-tuning of references for highly dynamic motions to account for model uncertainties.
In previous works \cite{chignoli2021online, song2022optimal} that perform the in-place barrel roll, nontrivial tuning is required to account for model mismatch, and the robot has to be locked to a pre-defined configuration before landing. By comparison in this work, the landing is optimized online, and the model mismatch is accounted for by {\sc Cafe-Mpc}.

\subsubsection{Comparison To Template MPC}
To further investigate the capability of {\sc Cafe-Mpc}, we compare it with a conventional MPC approach that plans over simplified models. We use the Hybrid Kinodynamic MPC (HKD-MPC) developed in our previous work \cite{li2022versatile}. The HKD-MPC reasons about SRB dynamics, and employs contact-dependent kinematics (full joint kinematics for swing legs and prismatic foot for stance legs). The HKD-MPC computes the CoM trajectory (translation and orientation), foot placements, and GRFs. The GRFs are directly used for stance leg control. A separate swing controller is used to track swing trajectories that interpolate the optimized foot placements. The HKD-MPC has been shown with robust performance to synthesize running jump, and strong disturbance rejection capability\cite{li2022versatile}. We utilize the already proficient HKD-MPC to follow the identical reference as {\sc Cafe-Mpc} in the task (1). Figure.~\ref{fig:br_hkdmpc} shows the time-series snapshots of the resulting motion in simulation. The robot behaves reasonably until after taking off, where the swing legs cross over the body fast, reducing the body angular momentum and resulting in an unsafe landing configuration. Fundamentally, the failure arises from the fact the HKD-MPC does not coordinate the body angular momentum and the leg angular momentum. By comparison, the {\sc Cafe-Mpc} explores the whole-body dynamics (though over a short prediction horizon) and accounts for the conservation of angular momentum implicitly via the whole-body dynamics for re-orienting the body for safe landing.

\subsubsection{Fosbury Flop}
\setcounter{figure}{14}
\begin{figure}
    \centering
    \includegraphics[width=0.9\linewidth]{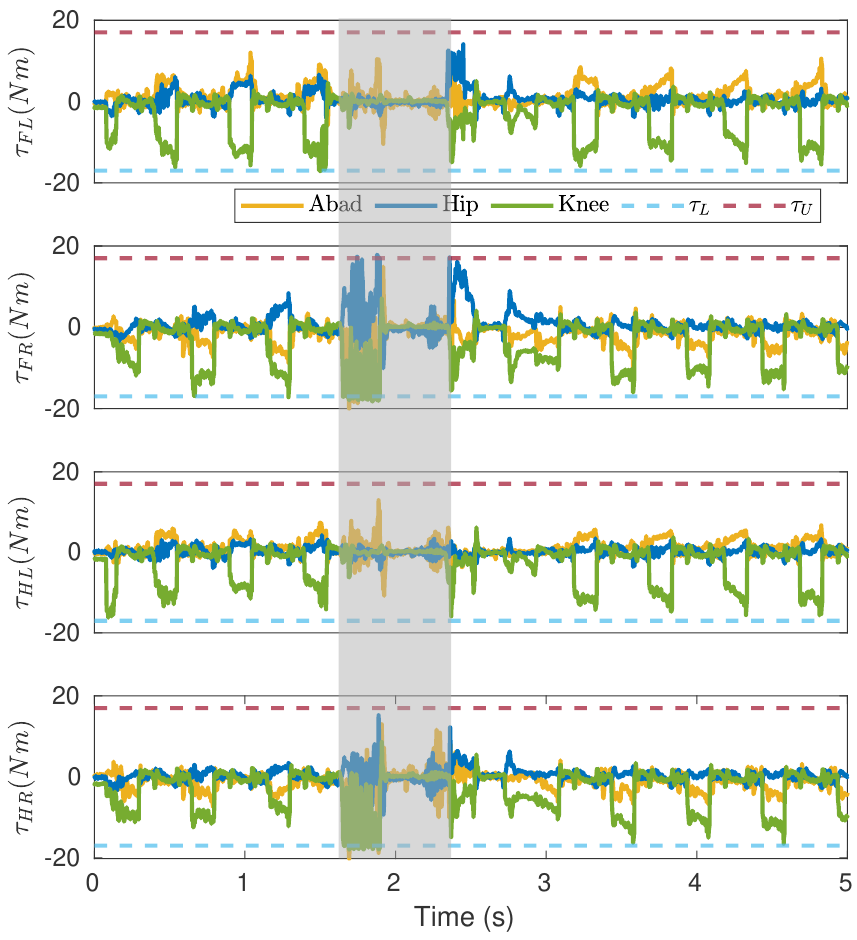}
    \caption{Forward velocity, height, and roll angle of the Mini Cheetah hardware during the mid-run barrel roll. The dashed line indicates the height of the string over which the robot jumps. (update this figure to indicate barrel roll)}
    \label{fig:flop_state}
\end{figure}
The Fosbury flop is a technique that a well-trained athlete utilizes in the high jump sport. We further push the capability of the proposed control framework to achieve this extremely dynamic and challenging motion on the robot hardware. We imitate the Fosbury Flop with a running barrel roll without an intermediate full stance. The robot alternates between left and right legs (pacing) at a certain speed. It then decelerates while holding the right legs at stance, and makes a barrel roll (Fobusry flop). Pacing is resumed with the same speed after landing and a hopping step. The composed trajectory is obtained by simply terminating the pacing reference with left legs at stance, starting the barrel roll reference with right legs at stance, and connecting them together without additional effort. Figure.~\ref{fig:online_smoothing} depicts the composed running barrel roll reference and the executed trajectory obtained in simulation. The green circle represents the discontinuity of the composed reference. The {\sc Cafe-Mpc} is capable of synthesizing a smooth transition online to deviate from this discontinuity. The jump in the actual state at the end of the barrel roll arises from the impact at touchdown. 
Figure~\ref{fig:flop_snapshots} shows time-series snapshots of the executed motion on the robot hardware. As shown in Fig.~\ref{fig:flop_state}, the robot jumps up to 0.52 m, and clears the string at 0.4 m high. The attained clearance height matches that of the MIT Cheetah 2 \cite{park2017high}, which is twice the size of the Mini Cheetah, though the jumping manners are different. The robot runs at up to 0.8 m/s before and after the barrel roll. Figure~\ref{fig:flop_torque} depicts the torque measurements, indicating that the torque limits are almost always satisfied with small violations at a few moments. We note that there are relatively large torque oscillations on the right stance legs at taking off. These oscillations happen at about 50 Hz, close to the {\sc Cafe-Mpc} update frequency. A hypothesis of these oscillations thus is that the control bandwidth of {\sc Cafe-Mpc} is not sufficient to resolve the fast dynamics response at taking off. Future work will investigate this problem in depth. However, that the MPC is able to recover, shows the capabilities of the online synthesis.

\begin{figure}
    \centering
    \includegraphics[width=0.9\linewidth]{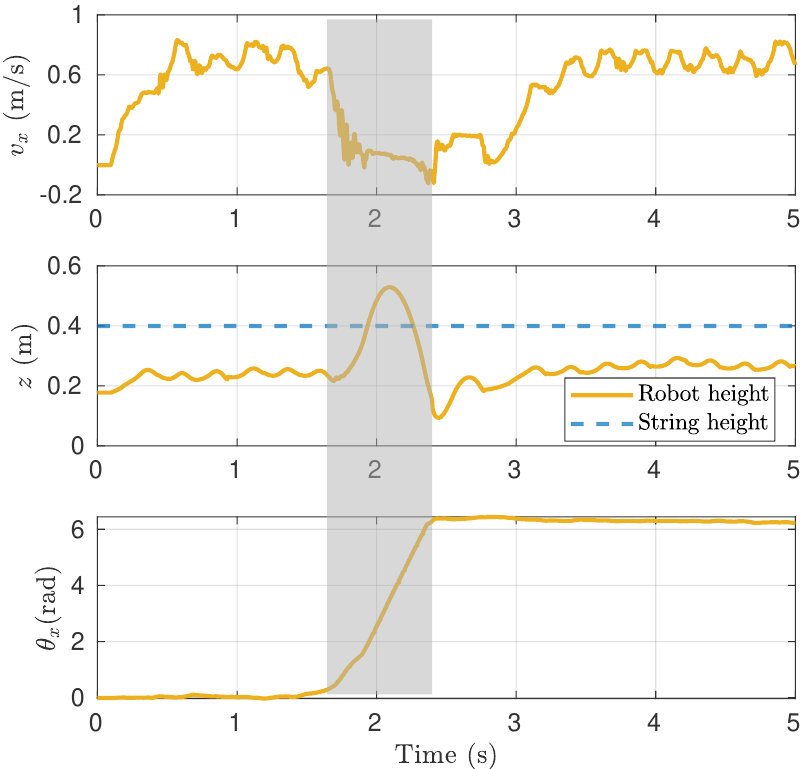}
    \caption{Joint torques of the Mini Cheetah hardware during the mid-run barrel roll. Dashed lines indicate torque limits. (update this figure to indicate barrel roll)}
    \label{fig:flop_torque}
\end{figure}

\begin{figure}
    \centering
    \includegraphics[width=0.95\linewidth]{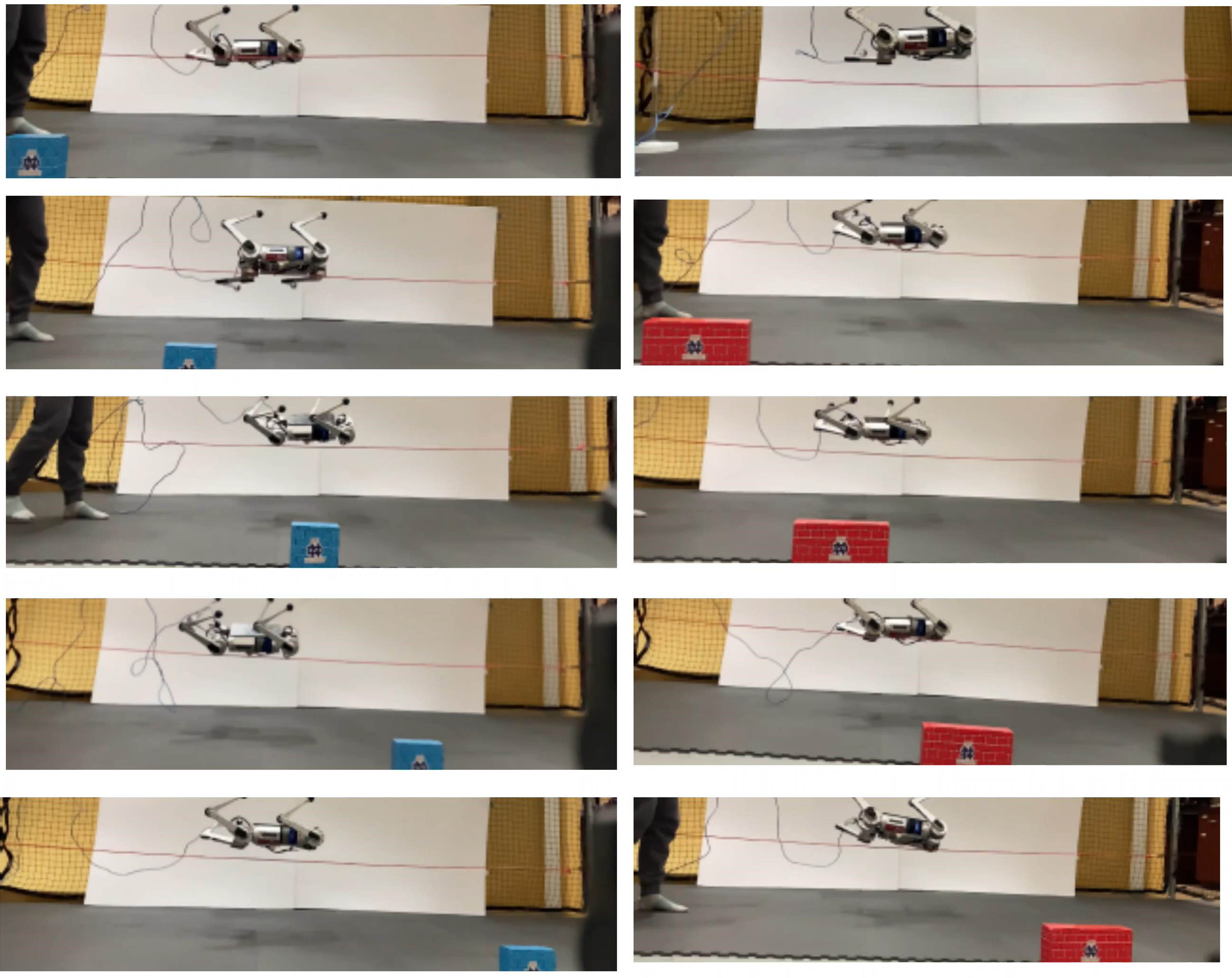}
    \caption{Reliability test of performing mid-run barrel roll on Mini Cheetah. 20 experiments were conducted with a success rate of $70\%$. 10 successful experiments are shown here, where the blue and red boxes at different locations indicate the experiments were performed independently.}
    \label{fig:reliability}
\end{figure}

\subsubsection{Reliability Test}
We repeat the Fosbury-Flop experiment 22 times to evaluate the reliability of the proposed control framework. Sixteen experiments are successful, resulting in a success rate of $72.6\%$. Figure.~\ref{fig:reliability} depicts screenshots of 10 successful experiments, where the blue and red boxes at different locations indicate these experiments were conducted independently. While two failed tests are due to low battery, and one due to temporary loss of Ethernet connection, the rest of the failed experiments are mainly caused by the sliding of the stance feet immediately before the barrel roll. This sliding arises from the oscillation (similar as observed in Fig.~\ref{fig:flop_torque}) of the stance legs that break the friction constraints. We believe once the oscillation problem is resolved, the reliability of the proposed control framework will be further improved.

%% file: MS_01/08_Conclusion.tex
\section{Conclusions and Future Work}
\label{sec:conclusion}

\subsection{Discussions and Future Works}
The results in Section~\ref{sec:result_mhpc} show the enhanced performance of \cmpc with the increased horizon of the SRB plan. This improvement is explained as follows. The value function obtained at the initial state of the SRB plan provides a proxy of the viability or tracking performance. When this value function is used as a terminal cost, the WB plan has a better idea of how the current decision would affect the long-term performance. This approach is similar to the finite-horizon LQR when the terminal cost is given by the value function of an infinite-horizon LQR. 

There are a few limitations with the \cmpc. The orientation is represented by Euler angles, which are notoriously known for the singularity problem. Future work would use quaternion \cite{jackson2021planning, li2021model}, and unlock more gymnastic maneuvers such as mid-run front flip. Self-collision is currently not considered, which can cause the robot to fall during leg crossing. Contact timing optimization is another line of future research to account for contact mismatch. Our previous work \cite{li2020hybrid} and current research on timing-free optimization show promise in this direction. Further, the computational efficiency of \cmpc can further be improved with the real-time iteration scheme \cite{diehl2005real}.

The VWBC disguises the Riccati feedback controller in a QP problem. Results in Section~\ref{sec:res_vwbc} demonstrate that the VWBC enjoys the benefits of Riccati feedback closing the control loop at a higher rate, while at the same time satisfying necessary constraints. This feature allows the high-level \cmpc to be solved at coarse precision. One limitation of VWBC is that the foot contacts are currently specified using a known schedule, which can cause instability problems in the presence of a large contact mismatch. Future work would seed VWBC with contact detection \cite{bledt2018contact}.

\subsection{Conclusion}
The major contribution of this work is the control pipeline of \cmpc + VWBC. The \cmpc enables flexible scheduling of multi-fidelity models, multi-resolution time steps, and relaxed constraints along the prediction horizon. When compared to the whole-body MPC on the regular locomotion tasks, the \cmpc was shown to achieve better tracking performance due to an appended template SRB plan that grows the prediction horizon. Further, this performance enhancement for \cmpc does not necessarily increase the computational time with the appropriate design of the SRB plan, which is beneficial for on-board computing with limited computational resources. Thanks to the multiple-shooting iLQR, the \cmpc computes for free a local Q-value function for the current state of the whole-body model. This local Q-value function measures the long-horizon effects of the current control perturbations on the resulting cost. The VWBC minimizes this value function so that the resulting solution is encouraged to stay near the optimal path discovered by MPC. As a result, the VWBC is free of additional tuning, which is an advantage over conventional OSC-QP. %

The proposed control pipeline is sufficiently powerful that it can synthesize highly dynamic and complex behaviors on the fly. This capability is demonstrated via performing a running barrel roll that mimics the human Fosbury Flop on the MIT Mini Cheetah. The reference for this motion is obtained by simply connecting an in-place barrel roll trajectory to a pacing gait, without sophisticated processing. The running barrel roll experiment was repeated 22 times on the Mini Cheetah hardware, and a 72.6$\%$ success rate was achieved. Most failures were caused by the sliding of the stance feet.  
A template MPC that reasons about a hybrid kinodynamics model is shown to fail to complete an in-place barrel roll in simulation. This failure is due to the omitted leg momentum, indicating the importance of incorporating whole-body dynamics for such highly dynamic motions. 
Beyond quadrupeds, the proposed control pipeline is very general and could be applied to many other robotic platforms. Further work will investigate its application to humanoid robots, in particular.

%% file: MS_01/Appendix_01.tex
\section*{Appendix: Offline TO Design of In-Place Barrel Roll} \label{sec:append}
The contact sequence for the in-place barrel roll is given by $\{$FS, FR-HR, FT, FS$\}$, meaning full-stance $\rightarrow$ right-stance $\rightarrow$ flight $\rightarrow$ full-stance. Further, we append a recovery step ($\{$FT, FS$\}$) to the in-place barrel roll, in case the stance phase after landing is not sufficient to balance the robot. This choice is motivated by the additional steps that are often taken by human gymnasts after landing to assist in balancing. The contact timings for the barrel roll are heuristically determined by mimicking that of a backflip in previous work \cite{katz2019mini}.

\begin{figure}
    \centering
    \includegraphics[width=0.95\linewidth]{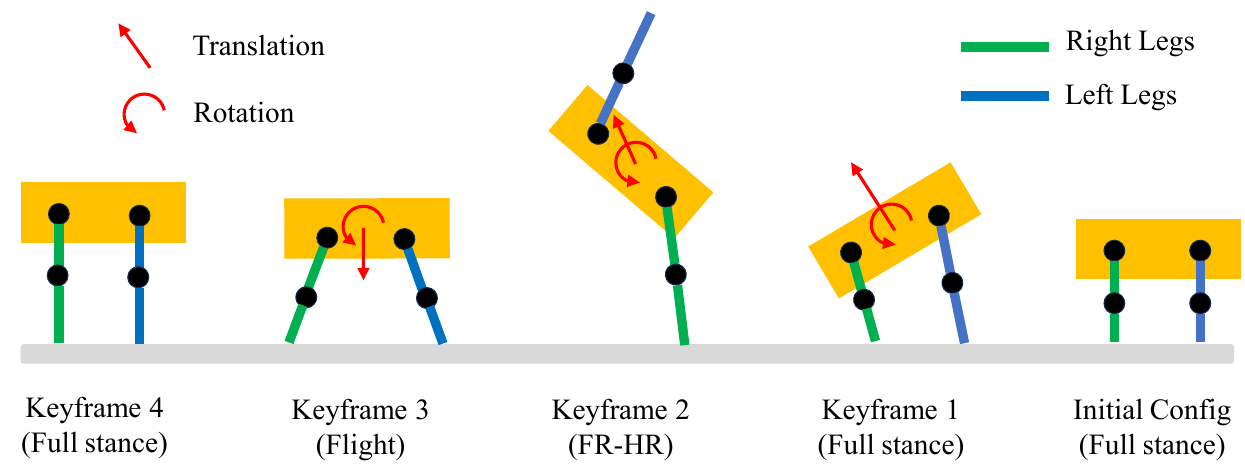}
    \caption{Illustration of keyframes used for barrel roll TO, depicted from the front view of the robot. Each keyframe is the desired terminal state for the associated phase (as labeled in parenthesis).}
    \label{fig:br_keyframes}
\end{figure}

\label{sec:br_reference}
The in-place barrel roll TO employs the whole-body dynamics. It shares almost the same problem structure as the whole-body planning problem of {\sc Cafe-Mpc} (Section~\ref{sec:wb_plan}), enabling tailoring available code for this specific task, and significantly reducing the amount of engineering efforts. There are two differences. First, we add to the in-place barrel roll TO a minimum body-height constraint at the full-stance phase after landing to avoid ground collision. Second, rather than tracking a reference trajectory, we manually design a few keyframes, one for each phase. The running cost and terminal cost of each phase are simply to track the allocated keyframes (Fig.~\ref{fig:br_keyframes}). The swing trajectory cost and the foot placement cost are not used in the barrel roll TO. 

The design of the keyframes is now discussed. We first design the floating-base states and then associate them with appropriate joint angles. The roll angle experiences a $2\pi$-rotation during the barrel roll. Assuming a constant roll rate, the roll rate is calculated by averaging $2\pi$ over the barrel roll duration. Given the contact timings, the roll angle at each keyframe is obtained with simple calculations. The lateral position ($y$) is similarly calculated with a constant lateral velocity ($v_y$) selected by the user. With the minimum landing height and flight duration, the vertical speed and position are determined assuming parabolic motion. Yaw and pitch angles are assumed to be always zero. An illustration of the keyframes is shown in Fig.~\ref{fig:br_keyframes}, which is drawn from the front view of the robot. For stance legs in keyframes 1 and 2, the joint angles are solved via inverse kinematics. For swing legs in keyframes 2 and 3, the joint angles are set to a reasonable landing configuration. Joint angles in keyframe 4 are determined via a default standing pose.

The resulting trajectory obtained from this offline TO is then used as a reference and an initial guess for the online {\sc Cafe-Mpc}. Though the offline TO is designed for the in-place barrel roll, the same trajectory is used to seed the running barrel roll online for multiple contact configurations. More specifically, the locomotion references Section~\ref{sec:loco_reference} are placed ahead and/or appended to the in-place barrel roll trajectory with simple manipulations, for instance, the roll angles of the appending trajectory are shifted by $2\pi$. Though there are no advanced techniques used here to glue the trajectories together, the {\sc Cafe-Mpc} plays a role of online synthesizing a smooth transition.